\documentclass[11pt]{article}

\usepackage[preprint]{acl} 

\usepackage{times}
\usepackage{latexsym}

\usepackage[T1]{fontenc}

\usepackage[utf8]{inputenc}

\usepackage{microtype}

\usepackage{inconsolata}

\usepackage{graphicx}

\usepackage[table]{xcolor}         

\usepackage{booktabs}       
\usepackage{amsmath}
\usepackage{amssymb}
\usepackage{mathtools}
\usepackage{amsthm}
\usepackage{basecommon} 

\colorlet{blueHighlightColor}{blue!05} 
\colorlet{wrongPredictionColor}{red!05} 
\colorlet{correctPredictionColor}{green!05} 
\definecolor{lightestgray}{gray}{0.95}
\colorlet{blueEmphasis}{blue!80}

\newcommand{\blueul}[1]{{\color{blueEmphasis}\underline{{\color{black}#1}}}}

\usepackage{caption}
\usepackage{listings}
\theoremstyle{plain}

\theoremstyle{definition}

\theoremstyle{remark}

\usepackage{algorithm}
\usepackage[noend]{algpseudocode}  
\algrenewcommand\alglinenumber[1]{\small #1:}

\title{Supervised Fine-tuning for Geometry-aware Instruction-following} 

\author{Allen Schmaltz \\
  Reexpress AI \\
  \texttt{allen@re.express} \\}

\begin{document}
\maketitle
\begin{abstract}
We examine a contrastive input encoding scheme for supervised fine-tuning of language models (LMs) that masks the content of incorrect generations up to a classification control token. This small-scale study, using a variation of the classic word ordering diagnostic task, suggests this straightforward approach generally preserves the generation accuracy of supervised fine-tuning that uses only positive examples, while increasing the admission rate of generations in the calibrated high-probability region of a \textsc{Similarity}-\textsc{Distance}-\textsc{Magnitude} ($\sdm$) estimator for binary classification of instruction-following. At the same time, we corroborate existing evidence that constructing a calibrated $\sdm$ estimator, which models the geometry of the feature-representation space, is not strictly dependent on specialized fine-tuning of the underlying LM's parameters.
\end{abstract}
\section{Introduction}

\textsc{Similarity}-\textsc{Distance}-\textsc{Magnitude} ($\sdm$) activations are a more robust and interpretable formulation of the standard softmax activation function \cite{Schmaltz-2026-SDM-Activations}. Their unique behavior is a result of instance-wise scaling terms derived from the geometry of the feature-representation space over the support set. When used as the final-layer activation over the frozen parameters of pre-trained language models (LMs), they are the basis for relatively robust \textit{estimators} of the predictive uncertainty for selective classification, as needed for conditional-branching decisions, for human-users as-well as for test-time search. With this newfound behavior, the focus of modeling shifts to maximizing the proportion of generations in the $\hrRegionFull$ ($\hrRegionShort$) region, a partitioning of the output distribution estimated to only contain points with class- and prediction-conditional accuracies at least $\alpha$, a value near 1 (here, $\alpha = 0.99$).

Calibrating an $\sdm$ estimator for binary classification of instruction-following of an LM assumes the hidden states taken as input to the $\sdm$ activation are fixed, with the calibration thresholds determined on a held-out set disjoint from the data used to learn the weights of the underlying network. Being able to construct a robust estimator without having to update the LM's parameters is a desirable property, since updating an LM's parameters is computationally expensive and risks degrading the existing generation capabilities of the LM. Nonetheless, in the case when we are otherwise updating the LM's parameters to improve generation quality, the question remains whether there are changes to the fine-tuning method that would be beneficial for the downstream, test-time $\sdm$ estimator. Relatedly, the interplay between joint generation-and-classification LM training in output (logit) space \citep[inter alia]{Schmaltz-Etal-2016-ClassificationAsTransduction,ZhangEtlAl-2025-Generative} and estimators that model the geometry of the feature-representation space remains underexplored.

Specifically, in the context of supervised fine-tuning (SFT), in this small-scale study on a single word ordering task, we seek to address the following targeted questions: \textit{Does adding contrastive verification supervision maintain the generation quality of positive sequences, and does it increase the proportion of generations that can be verified in the calibrated high-probability region of an $\sdm$ estimator of instruction-following?}

We find initial evidence that a contrastive input encoding scheme over negative and positive generations improves over standard supervised fine-tuning in increasing the proportion of generations in the $\hrRegionShort$ region at a fixed $\alpha$ without consistent degradation in generation quality for the task. At the same time, we corroborate existing work that constructing a calibrated $\sdm$ estimator, at least in the conservative sense, is not strictly dependent on the fine-tuning approach. A model fine-tuned with standard SFT over positive sequences, as well as the base model itself, can be calibrated for instruction-following with an $\sdm$ estimator.

\section{Setting}
Existing work has examined using $\sdm$ estimators for binary classification of instruction-following over closed-weight language models by using an ensemble of exogenous closed-weight and open-weight models to construct the estimate of the predictive uncertainty \citep{Schmaltz-2026-ReexpressMCPServer}. Here, we have access to the weights of the model, and we seek to construct an estimate of instruction-following using the model itself \textit{and} to improve the generation quality of that same model.

We assume each training example consists of a prompt and a ``positive'' generation, which we seek for the model to learn to generate, paired with a static (offline) corresponding ``negative'' generation, which we seek to discourage the model from generating. Unlike the typical setting of post-hoc preference fine-tuning \cite[inter alia]{SchulmanEtAl-2017-PPO,OuyangEtAl-2022-RLHF,Rafailov-EtAl-2023-DPO}, we seek to jointly train the LM for generation and classification \citep[inter alia]{Schmaltz-Etal-2016-ClassificationAsTransduction,ZhangEtlAl-2025-Generative}, here specifically binary classification of instruction-following to distinguish positive and negative generations. In the present small-scale study, for computational efficiency, we do not consider generating hard negatives during LM training, nor reinforcement learning approaches, comparisons to which we leave to future work. (We do, however, consider generated negatives for calibrating the test-time estimators, as discussed below.) To further minimize computational costs, we use LoRA fine-tuning \cite{HuEtAl-2022-LoRA-Fine-tuning}, rather than full-parameter fine-tuning.

Specifically, each example in the LM's fine-tuning dataset, $\trainLMSplit=\{(\vt_n, \vs^-_n, \vs^+_n)\}_{n=1}^{N}$, consists of an input prompt, $\vt$, which is a sequence of tokens, paired with positive (class 1), $\vs^+$, and negative (class 0), $\vs^-$, examples that are sequences that do and do not, respectively, follow the instructions of the given prompt. Each $\vs^+$ and $\vs^-$ sequence is assumed to end with special control tokens representing a positive or negative, respectively, binary classification label (see \S~\ref{sec:contrastive-masking}). The LM, $\pi$, generates a completion given a prompt, $\hat{\vs} \sim \pi(\vt)$. We assume analogous labeled datasets for training and calibrating test-time estimators. These splits, $\trainSplit$ and $\calibrationSplit$, respectively, are disjoint from $\trainLMSplit$.

To label generated responses, we assume access to a function, $r: (\vs^+, \hat{\vs}) \mapsto \{0,1\}$, which provides a binary estimate of whether the generation matches the reference positive sequence. In the present work, $r$ is deterministic (similar to ``verifiable rewards'' in reinforcement learning), in our case via exact string matches.

\section{Methods}

We seek to improve the generation quality of the model, as in standard SFT, \textit{and} construct a standard final-layer $\sdm$ activation layer calibrated for binary classification of instruction-following.\footnote{The present targeted study keeps this choice of estimator fixed and does not consider alternative calibration methods. See \citet{Schmaltz-2026-SDM-Activations} for a comparison of representative Frequentist, Bayesian, and empirically-motivated methods.} In this context, we examine a variation of the contrastive input encoding scheme of \citet{ZhangEtlAl-2025-Generative} and its impact on the test-time estimators that are calibrated once LM fine-tuning has completed.

\subsection{Encoding: $\encodingContrastiveMasking$}\label{sec:contrastive-masking}

Each sequence $\vs$ ends with special control tokens: In the present work, \texttt{<verified>Yes</verified>} for sequences that correctly follow the instructions in the prompt, and \texttt{<verified>No</verified>}, otherwise. We refer to this encoding scheme as $\encodingContrastiveMasking$ (CM), because it includes a binary mask for use when calculating the next-token loss during training. (See Table~\ref{tab:contrastive-encoding}.) In addition to typical masking of the prompt sequence and padding tokens (if applicable), we mask the loss for all tokens in negative sequences up to the final \texttt{<verified>No</verified>} control tokens. 

This encoding scheme builds on existing joint generation-and-classification work, with the key difference that CM is not intended as an end in itself; the $\sdm$ estimator is the geometry-aware verifier at test time.\footref{fn:sdm-force-decoded-tokens} Unlike \citet{Schmaltz-Etal-2016-ClassificationAsTransduction}, we do not generate explanations nor corrections, nor do we consider test-time search, in our simplified diagnostic setting. This masking scheme itself is effectively the same as used in the joint generation and verification objective of GenRM \citep{ZhangEtlAl-2025-Generative} with some minor implementation differences. Rather than using a tuned mixing coefficient between binary classification and the standard SFT next-token loss over positive sequences\footnote{See Eq. 4 (of the ICLR version) in \citet{ZhangEtlAl-2025-Generative}.}, we fix the relative weighting through the encoding and data composition, with all sequences receiving verification supervision, as detailed next. 

\subsection{CM Next-token Loss} 
The positive sequences are supervised as in standard supervised fine-tuning, but only the final verification tag sequences are supervised for negative sequences in the masked next-token loss:
\begin{align}\label{eq:sdm-next-token-loss}
\gL_{\rm{CM}} &= -\frac{1}{M} \sum_{n=1}^{N} \sum_{i=1}^{\lvert \vs_n \rvert} m_{n_i} \log_{e}\left(
\frac{
e^{z_{{n_i}, v_{n_i}}}
}{
\sum^{\lvert \gV \rvert}_{v=1}{e^{z_{{n_i}, v}}}
}
\right), \notag \\ 
M &= \sum_{n=1}^{N} \sum_{i=1}^{\lvert \vs_n \rvert} m_{n_i}
\end{align}
where $N$ is the count of response sequences in the mini-batch; $\vs_n$ is the $n^{\rm{th}}$ response sequence, which contains $\lvert \vs_n \rvert$ tokens; $v_{n_i}$ is the target token index at position $i$ in $\vs_n$ over the vocabulary, $\gV$; and $z_{{n_i}, v}$ is the output logit from the final linear layer of the LM (not an $\sdm$ activation) predicting that target token, conditioned on the preceding context.

$M $ counts unmasked tokens, over which the loss is averaged. Specifically, the binary mask $m_{n_i} \in \{ 0, 1 \}$ is 0 for all prompt tokens, and 1 for a positive response sequence, including its answer, formatting tags, \texttt{Yes} verification tag sequence, and assistant-termination tokens. The mask is 0 for negative responses, except for the complete \texttt{No} verification tag sequence, \texttt{<verified>No</verified>}, and as such, negative sequences remain in the conditioning context, but are not directly supervised.

\begin{table*}[h]

  \centering
  \resizebox{0.7\textwidth}{!}{%
  \begin{tabular}{l p{0.9\textwidth}}
    \toprule
 \textsc{Sequence type} & \textsc{Sequence tokens} \\
\midrule
Prompt ($\vt$) & \texttt{Complete the sentence `Neat plans' by reordering all of the following without adding new punctuation nor words: `fail luck. without'. Only reply with the sentence in the XML \textless sentence\textgreater{} \textless/sentence\textgreater{} followed by \textless verified\textgreater Yes\textless/verified\textgreater{} if your answer correctly addressed the instructions, and \textless verified\textgreater No\textless/verified\textgreater{} if it did not.}  \\
\midrule
Negative completion ($\vs^-$) & \texttt{\textless sentence\textgreater Neat plans without fail luck.\textless/sentence\textgreater\textbackslash n\blueul{\textless verified\textgreater No\textless/verified\textgreater}} \\
\midrule
Positive completion ($\vs^+$) & \texttt{\blueul{\textless sentence\textgreater Neat plans fail without luck.\textless/sentence\textgreater\textbackslash n\textless verified\textgreater Yes\textless/verified\textgreater}} \\
    \bottomrule
  \end{tabular}
  }
 
\caption{Illustrative example prompt and completions for the word ordering task using the $\encodingContrastiveMasking$ encoding. The next-token loss is only calculated over the \blueul{underlined} tokens.}
\label{tab:contrastive-encoding}
\end{table*} 

\subsection{Training-time $\sdm$ estimator}\label{sec:training-time-sdm-estimator}

 During fine-tuning, at the end of each epoch, holding the parameters of the underlying network fixed, we calibrate an $\sdm$ estimator for binary classification of instruction-following. This is a two-step process where we first train a final-layer $\sdm$ activation, which is a small adaptor, and then calibrate that adaptor using held-out data. The training and calibration data for this are disjoint from that of the LM parameter updates.
 
The $\sdm$ activation models the geometry of the feature-representation space by adding instance-wise terms to the standard $\softmax$, here specifically for binary classification of instruction-following:
\begin{align}\label{eq:sdmActivation}
\sdm(\vz')_i = \frac{
(2+q)^{\dVal \cdot z'_i}
}{
\sum^1_{c=0}{(2+q)^{\dVal \cdot z'_c}}
}, i \in \{0, 1\}
\end{align}
where $\q \in [0, \lvert \trainSplit \rvert]$ ($\Similarity$) is the count of consecutive (depth-wise) nearest matches (by $L^2$ distance) in $\trainSplit$ that are correctly predicted \textit{and} match $\hat{y}$ of the test instance. 
$\Distance$, $\dVal \in [0,1]$, is the minimum (across classes) 1-quantile of the $L^2$ distance to the nearest match ($\dNearest$) in $\trainSplit$ relative to that of $\calibrationSplit$.
The $\Magnitude$, $z'$, is from the output logits of the adaptor, which is a 1-D CNN ($g: \vh \in \reals^D \mapsto \vh' \in \reals^M$) followed by a linear output layer ($\vh' \in \reals^M \mapsto \reals^2$). The vector $\vh'$ is used for $L^2$ dense matching to calculate $\q$ and $\dVal$.

The prediction-conditional estimate, ${\sdm(\vz')_{\hat{y}}}$, from the output of the activation layer is not used directly as the estimate of the predictive uncertainty.\footnote{The $\sdm$ estimator targets a different quantity, but it is also more robust, in general, to covariate shifts than directly using ${\sdm(\vz')_{\hat{y}}}$, which could still be overconfident for moderate-to-low $\q$ and $\dVal$ if $\vz'$ spikes over unusual data. The $\sdm$ estimator gains additional robustness by further partitioning the distribution via the rescaled $\Similarity$ ($\rescaledSimilarity$), since $\q$ itself is not directly dependent on the relative margin of $\vz'$.} Instead, after the $\sdm$ activation is trained, a separate calibration process constructs the $\sdm$ \textit{estimator}, which produces a class- and prediction-conditional estimate. Toward this end, $\q$ is first rescaled:
\begin{equation}\label{equation:rescaled-similarity} 
\rescaledSimilarity = \min \left ( \q, (2+q)^{\sdm(\vz')_{\hat{y}}} \right )
\end{equation}
Next, Alg.~2 from \citet{Schmaltz-2026-SDM-Activations} is run over a  held-out calibration set to find $\minRescaledSimiliarityForHRRegion$ and class-wise output thresholds $ [\psi_0, \psi_1]$ such that Eq.~\ref{eq:index-conditional-sdm-estimator} partitions the distribution of $\calibrationSplit$ into a $\hrRegionFull$ ($\sdmHR$) region for which the class- and prediction-conditional accuracy is at least $\alpha$: 
\begin{align}\label{eq:index-conditional-sdm-estimator}
\sdmHR := 
\begin{cases}
  \hat{y} & \text{if~}  \rescaledSimilarity \ge \minRescaledSimiliarityForHRRegion  \wedge \sdm(\vz')_{\hat{y}} \ge \psi_{\hat{y}}  \\
  \bot & \text{otherwise}
\end{cases}
\end{align}
where $\bot$ indicates a rejected (non-admitted) point. The prediction $\hat{y} = \argmax{\vz'}$ is from the $\sdm$ activation and not from the final linear layer of the underlying LM.

\paragraph{Input.}  Here, the $\sdm$ layer takes as input $\vt$ and $\vs$ up to and including the first \texttt{<verified>} tag.\footnote{\label{fn:sdm-force-decoded-tokens}The remaining \texttt{Yes</verified>} or \texttt{No</verified>} tokens are intended to be force-decoded based on the $\sdm$ estimator. This convention is for end-users, as well as for test-time search operations that continue conditionally generating, the latter of which we leave to future work.} The input vector $\vh$ is the concatenation of the final-layer hidden state of the final \texttt{<verified>} sequence position with the mean over the non-padding final-layer hidden states of the prompt and response prefix up to and including that boundary position.

\paragraph{LM epoch selection criteria.}

At the conclusion of each epoch, we train and calibrate an $\sdm$ estimator on a held-out set, disjoint from the LM's training set. We then generate one response for each prompt of a held-out validation set. (Here and elsewhere, we use greedy generation.) We take as our epoch selection criteria the epoch with the maximum true positive admitted proportion, defined as the count of responses with $y=\hat{y}=1$ in any nested region with $\alpha\ge0.95$, divided by all validation prompts.\footnote{This threshold is less stringent than the final target $\hrRegionShort$ region to avoid empty selection counts.}

\subsection{Test-time $\sdm$ estimator}

The $\sdm$ estimators used during LM parameter fine-tuning are discarded once training completes; they are only used in service of choosing the LM epoch. Once training has concluded, a separate $\sdm$ estimator is trained and calibrated over a dataset disjoint from the epoch selection data and the LM's training set. This data consists of positive sequences combined with generated hard negatives, falling back to static negative response sequences if needed to maintain class balance. This then is the $\sdm$ estimator used at test-time.

\section{Experiments}
\paragraph{Task.} We seek a representative task for which the base LM performs very poorly; can be objectively and quickly evaluated with minimal irreducible error; and requires more abstract skills to solve, such as search and constraint satisfaction, rather than purely static knowledge acquisition. A variation of the classic word ordering task \cite[inter alia]{Elman-1990-FindingStructureInTime,BrownEtal-1990-SMT,Brew-1992-Shake-and-Bake-MT,Schmaltz-EtAl-2016-WordOrderingWithoutSyntax}, in which the LM is tasked with reordering the final three words of a sentence, fulfills these desiderata (Table~\ref{tab:contrastive-encoding}).

\paragraph{Data.}

\begin{table}[!t]
  \centering
  \resizebox{\columnwidth}{!}{%
  \begin{tabular}{l l c c c}
    \toprule
   & & \textsc{Sentence} & \textsc{Exact Match} & \textsc{Verification} \\ 
    Dataset   & Model & \textsc{Acc.} & \textsc{Acc.} & \textsc{Balanced Acc.} \\
  \midrule 
$\datasetLMOrderHarvard$ & $\modelPhiThreeFiveInstruct$ & 0.42 & 0.42 & 0.55 \\
$\datasetLMOrderHarvard$ & $\modelPhiThreeFiveInstructFinetuned$ & 0.94 & - & - \\
$\datasetLMOrderHarvard$ & $\modelPhiThreeFiveInstructFinetunedAdditionalRun$ & 0.94 & - & - \\
$\datasetLMOrderHarvard$ & $\modelPhiThreeFiveInstructFinetunedCM$ & 0.93 & 0.93 & 0.71 \\
$\datasetLMOrderHarvard$ & $\modelPhiThreeFiveInstructFinetunedCMAdditionalRun$ & 0.93 & 0.92 & 0.78 \\

\midrule
$\datasetLMOrderHarvard$ & $\modelGemmaFourEtwoBInstruct$ & 0.0 & 0.0 & - \\
$\datasetLMOrderHarvard$ & $\modelGemmaFourEtwoBInstructFinetuned$ & 0.94 & - & - \\
$\datasetLMOrderHarvard$ & $\modelGemmaFourEtwoBInstructFinetunedAdditionalRun$ & 0.94 & - & - \\
$\datasetLMOrderHarvard$ & $\modelGemmaFourEtwoBInstructFinetunedCM$ & 0.96 & 0.95 & 0.68 \\
$\datasetLMOrderHarvard$ & $\modelGemmaFourEtwoBInstructFinetunedCMAdditionalRun$ & 0.94 & 0.93 & 0.72 \\

\midrule
$\datasetLMOrderWikipedia$ & $\modelPhiThreeFiveInstruct$ & 0.27 & 0.27 & 0.52 \\
$\datasetLMOrderWikipedia$ & $\modelPhiThreeFiveInstructFinetuned$ & 0.92 & - & - \\
$\datasetLMOrderWikipedia$ & $\modelPhiThreeFiveInstructFinetunedAdditionalRun$ & 0.91 & - & - \\
$\datasetLMOrderWikipedia$ & $\modelPhiThreeFiveInstructFinetunedCM$ & 0.93 & 0.93 & 0.70 \\
$\datasetLMOrderWikipedia$ & $\modelPhiThreeFiveInstructFinetunedCMAdditionalRun$ & 0.90 & 0.89 & 0.75 \\

\midrule
$\datasetLMOrderWikipedia$ & $\modelGemmaFourEtwoBInstruct$ & 0.0020 & 0.0020 & 0.53 \\
$\datasetLMOrderWikipedia$ & $\modelGemmaFourEtwoBInstructFinetuned$ & 0.90 & - & - \\
$\datasetLMOrderWikipedia$ & $\modelGemmaFourEtwoBInstructFinetunedAdditionalRun$ & 0.94 & - & - \\
$\datasetLMOrderWikipedia$ & $\modelGemmaFourEtwoBInstructFinetunedCM$ & 0.94 & 0.93 & 0.70 \\
$\datasetLMOrderWikipedia$ & $\modelGemmaFourEtwoBInstructFinetunedCMAdditionalRun$ & 0.93 & 0.92 & 0.68 \\

\midrule
$\datasetLMOrderWikipediaCVS$ & $\modelPhiThreeFiveInstruct$ & 0.15 & 0.15 & 0.50 \\
$\datasetLMOrderWikipediaCVS$ & $\modelPhiThreeFiveInstructFinetuned$ & 0.89 & - & - \\
$\datasetLMOrderWikipediaCVS$ & $\modelPhiThreeFiveInstructFinetunedAdditionalRun$ & 0.88 & - & - \\
$\datasetLMOrderWikipediaCVS$ & $\modelPhiThreeFiveInstructFinetunedCM$ & 0.91 & 0.90 & 0.66 \\
$\datasetLMOrderWikipediaCVS$ & $\modelPhiThreeFiveInstructFinetunedCMAdditionalRun$ & 0.87 & 0.86 & 0.70 \\

\midrule
$\datasetLMOrderWikipediaCVS$ & $\modelGemmaFourEtwoBInstruct$ & 0.0015 & 0.0015 & 0.54 \\
$\datasetLMOrderWikipediaCVS$ & $\modelGemmaFourEtwoBInstructFinetuned$ & 0.86 & - & - \\
$\datasetLMOrderWikipediaCVS$ & $\modelGemmaFourEtwoBInstructFinetunedAdditionalRun$ & 0.89 & - & - \\
$\datasetLMOrderWikipediaCVS$ & $\modelGemmaFourEtwoBInstructFinetunedCM$ & 0.91 & 0.90 & 0.64 \\
$\datasetLMOrderWikipediaCVS$ & $\modelGemmaFourEtwoBInstructFinetunedCMAdditionalRun$ & 0.91 & 0.90 & 0.60 \\
    \bottomrule
  \end{tabular}
  }  
    \caption{Generation results. \textsc{Exact Match Accuracy} only counts verbatim matches to the ground-truth (sentences, classifications, and formatting) as correct. \textsc{Sentence Accuracy} only considers the sentence content, after parsing the first available tagged sentence. \textsc{Verification Balanced Accuracy} is the mean of the class-conditional accuracies of the first generated Yes/No verdict by the LM (not the $\sdm$ activation) and is undefined if either true class is absent. The \textsc{+FT} models are not prompted for a classification, so the rightmost columns are omitted.} 
    \label{tab:experiments-generation-results}
\end{table}

We sample sentences from the 7.83 million line subset of processed English Wikipedia in the SentenceTransformers repository\footnote{\url{https://huggingface.co/datasets/sentence-transformers/wikipedia-en-sentences}} \cite{ReimersAndGurevych-2019-SentenceTransformers} restricted to sentences of length 5 to 60 words. (The mean length is approximately 19 words.) Static (offline) negatives are constructed by randomly permuting the suffix to not match the reference sentence. $\trainLMSplit$ consists of 5,000 underlying prompts, from which 5,000 static negatives are also constructed. 

The estimator for LM epoch selection uses 5,000 training and 5,000 calibration examples, disjoint from $\trainLMSplit$. After each epoch, we generate a response to every prompt in both splits before imposing class balance. Within each split, we keep up to 2,500 usable generated errors; use static negatives for any remaining negative examples; and use reference positives for the remaining 2,500 problems. In this way, each problem contributes one response. Correctness labels are taken as $y=r(\vs^+, \hat{\vs})$, which is independent of the generated classification token. A separate set of 1,000 validation prompts is used for epoch selection (\S~\ref{sec:training-time-sdm-estimator}).

\begin{table}[!t]
  \centering
  \resizebox{\columnwidth}{!}{%
  \begin{tabular}{l l  c c }
    \toprule
    Dataset   & Model & \textsc{Acc.} & $\frac{n}{|\testSplit|}$ \\
  \midrule 
$\datasetLMOrderHarvard$ & $\modelPhiThreeFiveInstruct$ & \allRejected & 0.0 \\
$\datasetLMOrderHarvard$ & $\modelPhiThreeFiveInstructFinetuned$ & \allRejected & 0.0 \\
$\datasetLMOrderHarvard$ & $\modelPhiThreeFiveInstructFinetunedAdditionalRun$ & \allRejected & 0.0 \\
$\datasetLMOrderHarvard$ & $\modelPhiThreeFiveInstructFinetunedCM$ & 0.94 & 0.11 \\
$\datasetLMOrderHarvard$ & $\modelPhiThreeFiveInstructFinetunedCMAdditionalRun$ & 1.0 & 0.26 \\

\midrule
$\datasetLMOrderHarvard$ & $\modelGemmaFourEtwoBInstruct$ & 1.0 & 0.63 \\
$\datasetLMOrderHarvard$ & $\modelGemmaFourEtwoBInstructFinetuned$ & \allRejected & 0.0 \\
$\datasetLMOrderHarvard$ & $\modelGemmaFourEtwoBInstructFinetunedAdditionalRun$ & \allRejected & 0.0 \\
$\datasetLMOrderHarvard$ & $\modelGemmaFourEtwoBInstructFinetunedCM$ & 1.0 & 0.037 \\
$\datasetLMOrderHarvard$ & $\modelGemmaFourEtwoBInstructFinetunedCMAdditionalRun$ & 1.0 & 0.10 \\

\midrule
$\datasetLMOrderWikipedia$ & $\modelPhiThreeFiveInstruct$ & \allRejected & 0.0 \\
$\datasetLMOrderWikipedia$ & $\modelPhiThreeFiveInstructFinetuned$ & 1.0 & 0.00042 \\
$\datasetLMOrderWikipedia$ & $\modelPhiThreeFiveInstructFinetunedAdditionalRun$ & \allRejected & 0.0 \\
$\datasetLMOrderWikipedia$ & $\modelPhiThreeFiveInstructFinetunedCM$ & 1.0 & 0.092 \\
$\datasetLMOrderWikipedia$ & $\modelPhiThreeFiveInstructFinetunedCMAdditionalRun$ & 1.0 & 0.081 \\

\midrule
$\datasetLMOrderWikipedia$ & $\modelGemmaFourEtwoBInstruct$ & 1.0 & 0.63 \\
$\datasetLMOrderWikipedia$ & $\modelGemmaFourEtwoBInstructFinetuned$ & \allRejected & 0.0 \\
$\datasetLMOrderWikipedia$ & $\modelGemmaFourEtwoBInstructFinetunedAdditionalRun$ & \allRejected & 0.0 \\
$\datasetLMOrderWikipedia$ & $\modelGemmaFourEtwoBInstructFinetunedCM$ & 0.99 & 0.049 \\
$\datasetLMOrderWikipedia$ & $\modelGemmaFourEtwoBInstructFinetunedCMAdditionalRun$ & 1.0 & 0.036 \\

\midrule
$\datasetLMOrderWikipediaCVS$ & $\modelPhiThreeFiveInstruct$ & \allRejected & 0.0 \\
$\datasetLMOrderWikipediaCVS$ & $\modelPhiThreeFiveInstructFinetuned$ & \allRejected & 0.0 \\
$\datasetLMOrderWikipediaCVS$ & $\modelPhiThreeFiveInstructFinetunedAdditionalRun$ & \allRejected & 0.0 \\
$\datasetLMOrderWikipediaCVS$ & $\modelPhiThreeFiveInstructFinetunedCM$ & \allRejected & 0.0 \\
$\datasetLMOrderWikipediaCVS$ & $\modelPhiThreeFiveInstructFinetunedCMAdditionalRun$ & 1.0 & 0.00050 \\

\midrule
$\datasetLMOrderWikipediaCVS$ & $\modelGemmaFourEtwoBInstruct$ & 1.0 & 0.20 \\
$\datasetLMOrderWikipediaCVS$ & $\modelGemmaFourEtwoBInstructFinetuned$ & \allRejected & 0.0 \\
$\datasetLMOrderWikipediaCVS$ & $\modelGemmaFourEtwoBInstructFinetunedAdditionalRun$ & \allRejected & 0.0 \\
$\datasetLMOrderWikipediaCVS$ & $\modelGemmaFourEtwoBInstructFinetunedCM$ & 1.0 & 0.0024 \\
$\datasetLMOrderWikipediaCVS$ & $\modelGemmaFourEtwoBInstructFinetunedCMAdditionalRun$ & 1.0 & 0.00017 \\

    \bottomrule
  \end{tabular}
  }  
    \caption{Selective classification results for instruction-following (content and formatting), among generations in the $\hrRegionShort$ region, with $\alpha=0.99$. A higher proportion of generations in the $\hrRegionShort$ region, $\frac{n}{|\testSplit|}$, is preferable, provided the accuracy is $\ge \alpha=0.99$. Results are averaged over 6 independently trained and calibrated $\sdm$ estimators. \allRejected\ denotes an empty region across all six estimators.} 
    \label{tab:experiments-verification-results-abbreviated} 
\end{table}

The test-time $\sdm$ estimators are similarly constructed using generated negatives with fallback to static examples as needed, but over separate disjoint sets containing 20,000 training examples and 20,000 calibration examples, where again the positive and negatives sequences are balanced by class.

At test time, the LM generates a completion from the prompt, and the test-time $\sdm$ estimator evaluates binary classification of instruction-following over the result. The in-distribution test set ($\datasetLMOrderWikipedia$) consists of 2,000 sentences.\footnote{We use the separate label $\datasetLMOrderWikipediaCalibrationSet$ for the calibration data of the test-time estimators.} We also consider a covariate shifted test set ($\datasetLMOrderWikipediaCVS$) of 2,000 sentences all of length 60, and an in-distribution test set consisting of the 720 classic IEEE ``Harvard'' sentences\footnote{\url{https://www.cs.columbia.edu/~hgs/audio/harvard.html}} ($\datasetLMOrderHarvard$) \cite{IEEE-Classic-Harvard-Speach-Quality-Sentences-1969}.

The distribution of positive and negative response sequences is model dependent in the main evaluation sets, reflecting the real-world setting of instruction-following of an LM used for generation. We additionally consider a control in which calibration and evaluation of the $\sdm$ estimators use static negatives and positives. We use the labels $\datasetLMOrderStaticWikipedia$ (which has $4,000$ sequences), $\datasetLMOrderStaticWikipediaCVS$ ($4,000$ sequences), and $\datasetLMOrderStaticHarvard$ ($1,440$ sequences) for these corresponding sets, where evaluation includes both the reference positive and reference static negative for each prompt. The corresponding $\sdm$ estimators are trained and calibrated with 20,000 static training and 20,000 static calibration examples.

\paragraph{Models.}

As base models, we use the \texttt{Phi-3.5-mini-instruct} model ($\modelPhiThreeFiveInstruct$) and the \texttt{Gemma-4-E2B-it} model ($\modelGemmaFourEtwoBInstruct$), which are decoder-only Transformer-based language model with 3.8 billion parameters  \citep{Abdin-2024-Phi3-TechReport} and 5.1 billion parameters\footnote{Intended for edge compute, the model has 2.3 billion parameters when excluding the per-layer embeddings.} \citep{Google-GemmaTeam-2026-Gemma4-TechReport}, respectively. These representative LMs are at a scale that enables fine-tuning, given available resources, for our controlled study. The parameters of the $\sdm$ activation layers generally follow those of previous works, but we use the nested estimator construction \citep[Appendix A.10]{Schmaltz-2026-SDM-Activations} with a resolution of $r=0.01$.

\paragraph{Comparisons.} The baselines are the models prior to fine-tuning ($\modelPhiThreeFiveInstruct$ and $\modelGemmaFourEtwoBInstruct$). We then compare to fine-tuning with the standard cross-entropy loss using only positive examples ($\modelPhiThreeFiveInstructFinetuned$ and $\modelGemmaFourEtwoBInstructFinetuned$), and fine-tuning using the $\encodingContrastiveMasking$ encoding ($\modelPhiThreeFiveInstructFinetunedCM$ and $\modelGemmaFourEtwoBInstructFinetunedCM$). In both sets of experiments across models, we also run LoRA fine-tuning a second time with a new seed, for which we use the subscript label ``2nd''. For reference, test-time $\sdm$ estimators are also constructed over the base models, using generated responses from the models prior to fine-tuning.

\section{Results}

\begin{table}[!t]
  \centering
  \resizebox{\columnwidth}{!}{%
  \begin{tabular}{l l  c c }
    \toprule
    Dataset   & Model & \textsc{Acc.} & $\frac{n}{|\testSplit|}$ \\
  \midrule 
$\datasetLMOrderStaticHarvard$ & $\modelPhiThreeFiveInstruct$ & 1.0 & 0.66 \\
$\datasetLMOrderStaticHarvard$ & $\modelPhiThreeFiveInstructFinetuned$ & 1.0 & 0.79 \\
$\datasetLMOrderStaticHarvard$ & $\modelPhiThreeFiveInstructFinetunedAdditionalRun$ & 1.0 & 0.73 \\
$\datasetLMOrderStaticHarvard$ & $\modelPhiThreeFiveInstructFinetunedCM$ & 1.0 & 0.76 \\
$\datasetLMOrderStaticHarvard$ & $\modelPhiThreeFiveInstructFinetunedCMAdditionalRun$ & 1.0 & 0.79 \\

\midrule
$\datasetLMOrderStaticHarvard$ & $\modelGemmaFourEtwoBInstruct$ & 1.0 & 0.0022 \\
$\datasetLMOrderStaticHarvard$ & $\modelGemmaFourEtwoBInstructFinetuned$ & 1.0 & 0.40 \\
$\datasetLMOrderStaticHarvard$ & $\modelGemmaFourEtwoBInstructFinetunedAdditionalRun$ & 1.0 & 0.36 \\
$\datasetLMOrderStaticHarvard$ & $\modelGemmaFourEtwoBInstructFinetunedCM$ & 1.0 & 0.74 \\
$\datasetLMOrderStaticHarvard$ & $\modelGemmaFourEtwoBInstructFinetunedCMAdditionalRun$ & 1.0 & 0.86 \\

\midrule
$\datasetLMOrderStaticWikipedia$ & $\modelPhiThreeFiveInstruct$ & 1.0 & 0.52 \\
$\datasetLMOrderStaticWikipedia$ & $\modelPhiThreeFiveInstructFinetuned$ & 1.0 & 0.63 \\
$\datasetLMOrderStaticWikipedia$ & $\modelPhiThreeFiveInstructFinetunedAdditionalRun$ & 1.0 & 0.64 \\
$\datasetLMOrderStaticWikipedia$ & $\modelPhiThreeFiveInstructFinetunedCM$ & 1.0 & 0.72 \\
$\datasetLMOrderStaticWikipedia$ & $\modelPhiThreeFiveInstructFinetunedCMAdditionalRun$ & 1.0 & 0.78 \\

\midrule
$\datasetLMOrderStaticWikipedia$ & $\modelGemmaFourEtwoBInstruct$ & 1.0 & 0.0082 \\
$\datasetLMOrderStaticWikipedia$ & $\modelGemmaFourEtwoBInstructFinetuned$ & 1.0 & 0.41 \\
$\datasetLMOrderStaticWikipedia$ & $\modelGemmaFourEtwoBInstructFinetunedAdditionalRun$ & 1.0 & 0.33 \\
$\datasetLMOrderStaticWikipedia$ & $\modelGemmaFourEtwoBInstructFinetunedCM$ & 1.0 & 0.76 \\
$\datasetLMOrderStaticWikipedia$ & $\modelGemmaFourEtwoBInstructFinetunedCMAdditionalRun$ & 1.0 & 0.82 \\

\midrule
$\datasetLMOrderStaticWikipediaCVS$ & $\modelPhiThreeFiveInstruct$ & 1.0 & 0.098 \\
$\datasetLMOrderStaticWikipediaCVS$ & $\modelPhiThreeFiveInstructFinetuned$ & 1.0 & 0.080 \\
$\datasetLMOrderStaticWikipediaCVS$ & $\modelPhiThreeFiveInstructFinetunedAdditionalRun$ & 1.0 & 0.17 \\
$\datasetLMOrderStaticWikipediaCVS$ & $\modelPhiThreeFiveInstructFinetunedCM$ & 1.0 & 0.18 \\
$\datasetLMOrderStaticWikipediaCVS$ & $\modelPhiThreeFiveInstructFinetunedCMAdditionalRun$ & 1.0 & 0.36 \\

\midrule
$\datasetLMOrderStaticWikipediaCVS$ & $\modelGemmaFourEtwoBInstruct$ & \allRejected & 0.0 \\
$\datasetLMOrderStaticWikipediaCVS$ & $\modelGemmaFourEtwoBInstructFinetuned$ & 1.0 & 0.10 \\
$\datasetLMOrderStaticWikipediaCVS$ & $\modelGemmaFourEtwoBInstructFinetunedAdditionalRun$ & 1.0 & 0.0027 \\
$\datasetLMOrderStaticWikipediaCVS$ & $\modelGemmaFourEtwoBInstructFinetunedCM$ & 1.0 & 0.55 \\
$\datasetLMOrderStaticWikipediaCVS$ & $\modelGemmaFourEtwoBInstructFinetunedCMAdditionalRun$ & 1.0 & 0.41 \\

    \bottomrule
  \end{tabular}
  }  
    \caption{Selective classification results averaged over 6 independent $\sdm$ estimators trained and calibrated on \underline{static} positive and negative sequences, for the $\hrRegionShort$ region with $\alpha=0.99$. The evaluation sets contain the same static positive and negative sequences across models.} 
    \label{tab:experiments-verification-results-abbreviated-static-resolution01-hr99} 
\end{table}

A possible concern with learning to classify negative sequences in this simple way is that it would lead to a degradation in generation quality of the positive responses, since the content of the negative sequences still appears in the conditioning context, even if not directly supervised. We do not observe consistent evidence of such degradation here, as shown in Table~\ref{tab:experiments-generation-results}, which evaluates the sentence content of the generations without using the $\sdm$ layer. Across datasets, training with $\encodingContrastiveMasking$ (\textsc{+CM}) neither consistently improves nor degrades generation quality of the underlying LM.

In all cases, fine-tuning itself substantively improves generation quality. The base models are quite poor at this diagnostic task, frequently failing to correctly reorder and/or adding extraneous punctuation or words. LoRA fine-tuning results in sentence-level accuracies near or above 0.9 for most of the models and datasets.

On the other hand, the balanced marginal accuracy of the classification tags (right column, Table~\ref{tab:experiments-generation-results}) generated by the models is relatively low even for the \textsc{+CM} models, reflecting poor discrimination over generated negative sequences. Here, $\encodingContrastiveMasking$ is not intended to be used alone for instruction-following; the test-time $\sdm$ estimators add the constraints over the feature-representation space of the model, and class- and prediction-conditional estimates of the predictive uncertainty.\footnote{Not the focus of the present work, but the $\sdm$ estimators also enable uncertainty-aware interpretability-by-exemplar.} 

The $\sdm$ estimators over the \textsc{+CM} models generally have higher admission rates, $\frac{n}{|\testSplit|}$, in the $\hrRegionShort$ region than the models trained only with positive sequences (\textsc{+FT}), as seen in Table~\ref{tab:experiments-verification-results-abbreviated}, despite comparable generation quality.  (See the full Table~\ref{tab:experiments-verification-results} for the corresponding class- and prediction-conditional accuracies.) As shown in Table~\ref{tab:experiments-verification-results-calibration}, only two of 24 $\sdm$ estimators for the models trained with only positive sequences have a finite $\minRescaledSimiliarityForHRRegion$ at $\alpha=0.99$. This  admission rate trend persists when stratifying by lower probability regions, as shown in Table~\ref{tab:experiments-verification-results-resolution01nested95}, which shows the nested regions $\ge 0.95$. The conditional accuracies that do fall below the target threshold tend to be small counts lower than the proportion of the threshold. In the one borderline case of $\modelPhiThreeFiveInstructFinetunedCM$ over the $\datasetLMOrderHarvard$ dataset, the under-coverage in the class 1 prediction-conditional average is the result of 1 or 2 false positives across each estimator. Those errors, stemming from the same two responses, are arguably within the aleatoric (irreducible) error of the task (see Table~\ref{tab:experiments-verification-results}, caption). 

\begin{table*}
  \centering
  \setlength{\fboxsep}{1pt}
  \resizebox{1.0\textwidth}{!}{%
  \begin{tabular}{l l c c c c c c c c c}
    \toprule

    & &  \multicolumn{8}{c}{Class-conditional}  \\
    & & \multicolumn{1}{c}{$y=0$} & \multicolumn{1}{c}{$y=1$} &  \multicolumn{1}{c}{$y=0$} & \multicolumn{1}{c}{$y=1$} &  \multicolumn{1}{c}{$y=0$} & \multicolumn{1}{c}{$y=1$} &  \multicolumn{1}{c}{$y=0$} & \multicolumn{1}{c}{$y=1$} \\
    \cmidrule(r){3-4} \cmidrule(r){5-6} \cmidrule(r){7-8} \cmidrule(r){9-10}
    $\sdm$ Calibration set  & Model & $\sdm$ loss & $\sdm$ loss & \textsc{Acc.} & \textsc{Acc.} & \textsc{Mean} $\q$ & \textsc{Mean} $\q$ & $\psi_0$ & $\psi_1$ & $\minRescaledSimiliarityForHRRegion$\\
          
 \midrule
$\datasetLMOrderWikipediaCalibrationSet$ & $\modelPhiThreeFiveInstruct$ & 0.33 & 0.27 & 0.83 & 0.86 & 170 & 27 & N/A  & N/A & N/A \\
$\datasetLMOrderWikipediaCalibrationSet$ & $\modelPhiThreeFiveInstructFinetuned$ & 0.14 & 0.12 & 0.91 & 0.95 & 1400 & 150 & 0.99\textsuperscript{\dag} & 1.0\textsuperscript{\dag} & 1800\textsuperscript{\dag} \\
$\datasetLMOrderWikipediaCalibrationSet$ & $\modelPhiThreeFiveInstructFinetunedAdditionalRun$ & 0.13 & 0.11 & 0.92 & 0.95 & 1400 & 130 & N/A & N/A & N/A \\
$\datasetLMOrderWikipediaCalibrationSet$ & $\modelPhiThreeFiveInstructFinetunedCM$ & 0.078 & 0.066 & 0.95 & 0.97 & 1500 & 500 & 0.99 & 1.0 & 1400 \\
$\datasetLMOrderWikipediaCalibrationSet$ & $\modelPhiThreeFiveInstructFinetunedCMAdditionalRun$ & 0.086 & 0.075 & 0.94 & 0.97 & 1500 & 420 & 0.99 & 1.0 & 1400 \\

\midrule
$\datasetLMOrderWikipediaCalibrationSet$ & $\modelGemmaFourEtwoBInstruct$ & 0.049 & 0.037 & 0.98 & 0.98 & 1400 & 1200 & 0.99 & 1.0 & 1200 \\
$\datasetLMOrderWikipediaCalibrationSet$ & $\modelGemmaFourEtwoBInstructFinetuned$ & 0.19 & 0.14 & 0.88 & 0.94 & 1200 & 55 & N/A & N/A & N/A \\
$\datasetLMOrderWikipediaCalibrationSet$ & $\modelGemmaFourEtwoBInstructFinetunedAdditionalRun$ & 0.20 & 0.16 & 0.87 & 0.94 & 1300 & 50 & N/A & N/A & N/A \\
$\datasetLMOrderWikipediaCalibrationSet$ & $\modelGemmaFourEtwoBInstructFinetunedCM$ & 0.094 & 0.071 & 0.95 & 0.96 & 1500 & 340 & 0.99 & 0.99 & 1300 \\
$\datasetLMOrderWikipediaCalibrationSet$ & $\modelGemmaFourEtwoBInstructFinetunedCMAdditionalRun$ & 0.095 & 0.076 & 0.93 & 0.98 & 1500 & 240 & 0.99 & 1.0 & 1100 \\
         
    \bottomrule
  \end{tabular}
  }  
    \caption{Summary statistics over the calibration splits of the final-layer $\sdm$ estimators used at test-time, averaged over 6 independently calibrated $\sdm$ estimators. $\psi_0$ and $\psi_1$ are the class-wise output thresholds and $\minRescaledSimiliarityForHRRegion$ is the rescaled $\Similarity$ value used to determine the selection criteria introduced in \citet{Schmaltz-2026-SDM-Activations}, which determines the $\hrRegionFull$ region (here, with $\alpha=0.99$). $\sdm$ loss is the loss over the binary classification task from the $\sdm$ activation. N/A indicates that none of the six $\sdm$ estimators has an $\hrRegionShort$ region at $\alpha=0.99$. \textsuperscript{\dag}Threshold means for $\modelPhiThreeFiveInstructFinetuned$ use the two of six $\sdm$ estimators with an $\alpha=0.99$ $\hrRegionShort$ region.} %
  \label{tab:experiments-verification-results-calibration} 
\end{table*}

The relatively high admission rate of the base $\modelGemmaFourEtwoBInstruct$ model seen in the marginal results of Table~\ref{tab:experiments-verification-results-abbreviated} reflects confident verification of that LM's frequent generation mistakes, as opposed to a high true positive admitted proportion. (As seen in Table~\ref{tab:experiments-generation-results}, this base model very rarely generates a response sequence that correctly follows the prompt's instructions.) The partitioning of an $\sdm$ estimator will reflect the data distribution of the calibration set, so to reduce possible confounding from the distribution of model generated responses, Table~\ref{tab:experiments-verification-results-abbreviated-static-resolution01-hr99}  (full Table~\ref{tab:experiments-verification-results-static}) refits and evaluates six $\sdm$ estimators only on static positive and negative sequences. The general trend of the \textsc{+CM} models having higher admission rates in the $\alpha=0.99$ region than \textsc{+FT} models persists in this controlled setting. In this way, a difference in overall generation quality, or the distribution of the test-time hard negatives, does not in itself explain the differences in admission in the $\hrRegionShort$ region.

Within the results for each individual model, the difference in admission rates across datasets is consistent with previous works showing that $\sdm$ estimators tend to be more conservative as the test-time data deviates from the distribution of the calibration data. Across models, the covariate-shifted $\datasetLMOrderWikipediaCVS$ dataset tends to have lower admission rates than the in-distribution datasets. 

\section{Discussion}

Specialized fine-tuning of an LM's parameters is not strictly required to construct a calibrated test-time $\sdm$ estimator. These experiments add to the growing evidence that an $\sdm$ estimator calibrated over the frozen weights of the model, regardless of how it was trained, will tend to reflect the predictive uncertainty given the provided reference class, as represented by the data used to calibrate the $\sdm$ estimator. This pattern also holds for the base models prior to fine-tuning, where the very low task accuracies of the base models are reflected in commensurately more conservative high-probability regions. This has important implications for evaluating models for real applications, shifting the focus of evaluation from generation quality alone to assessing the proportion of well-calibrated sequences that can be generated \textit{and} reliably verified with high-probability.

\section{Conclusion}

We examined a straightforward contrastive encoding scheme that can increase the proportion of generations that can be verified in the calibrated high-probability region of a final-layer $\sdm$ estimator. An $\sdm$ estimator for instruction-following can be constructed without specialized fine-tuning of the underlying LM's parameters, and in the conservative sense, without fine-tuning itself, but we provide initial evidence of increased admission rates when using this input encoding compared to fine-tuning on positive sequences alone.


\section*{Limitations}

These initial results are of practical interest given that they can be readily applied to existing LM fine-tuning pipelines. Architectural changes that would require a full re-training are not needed for contemporary LMs that have already been trained using significant resources. However, for at least some applications with existing LMs, fine-tuning itself may not be necessary to achieve acceptable abstention rates (for a given $\alpha$, as needed for a particular task) and is typically not the first option to consider given the computational costs. In particular, a targeted increase in the size of $\trainSplit$ and $\calibrationSplit$ of the $\sdm$ estimator is typically the first option to consider, and the interpretability-by-exemplar capabilities of the $\sdm$ estimator can be used to aid such targeted data collection.

We hypothesize that test-time search strategies can be learned without changing the loss or basic setup beyond the data, enabling robust, uncertainty-aware conditional branching. In principle, generation can continue, or the conversation can be altogether reset with the negative sequence as part of the prompt, if a \texttt{<verified>No</verified>} control sequence is encountered. As noted in the main text, the final \texttt{Yes</verified>} or \texttt{No</verified>} tokens are intended to be force-decoded using the output of the $\sdm$ estimator; thus, these branching decisions would be made in the joint representation and output space of the $\sdm$ estimator, rather than the output (``token'') space of typical existing test-time search strategies.\footnote{In the present work, we only consider binary verification tags. For test-time search, it may be helpful to also include a tag for when the $\sdm$ estimator is uncertain, while otherwise retaining binary modeling for the $\sdm$ estimator.} Relatedly, we hypothesize that training with hard negatives generated online during training would be beneficial, at the expense of additional training compute. We leave these to future higher-resourced experiments for controlled examination across models.

Even though the $\encodingContrastiveMasking$ encoding scheme masks the content of the negative sequences, if a static negative is in the far tail of the output distribution of the underlying language model $\pi$, the probability of generating the negative sequence could potentially rise (even if still low in absolute terms) relative to never fine-tuning with such a sequence. The $\sdm$ estimator constrains negative generations commensurate to the chosen $\alpha$, but in the absence of a test-time $\sdm$ estimator, the test-time constraint would be the less robust estimate encoded in the final control sequences, which would be similar to using a $\softmax$ activation as an estimator of the predictive uncertainty. For this reason, we do not recommend using $\encodingContrastiveMasking$ without a test-time $\sdm$ estimator. Similarly, static negative examples in the tail of the output distribution of $\pi$ with very high-risk content are a priori better suited for the training and calibration sets of the test-time $\sdm$ estimator trained over the frozen parameters of $\pi$. 

This initial study only covers a single diagnostic task, two small models, and a small number of training runs. We also have not examined the relative impact on the generation quality for other tasks.

\bibliography{sdm_lm}

\begin{thebibliography}{19}
\providecommand{\natexlab}[1]{#1}

\bibitem[{Abdin et~al.(2024)Abdin, Aneja, Awadalla, Awadallah, Awan, Bach,
  Bahree, Bakhtiari, Bao, Behl, Benhaim, Bilenko, Bjorck, Bubeck, Cai, Cai,
  Chaudhary, Chen, Chen, Chen, Chen, Chen, Cheng, Chopra, Dai, Dixon, Eldan,
  Fragoso, Gao, Gao, Gao, Garg, Giorno, Goswami, Gunasekar, Haider, Hao,
  Hewett, Hu, Huynh, Iter, Jacobs, Javaheripi, Jin, Karampatziakis, Kauffmann,
  Khademi, Kim, Kim, Kurilenko, Lee, Lee, Li, Li, Liang, Liden, Lin, Lin, Liu,
  Liu, Liu, Liu, Liu, Luo, Madan, Mahmoudzadeh, Majercak, Mazzola, Mendes,
  Mitra, Modi, Nguyen, Norick, Patra, Perez-Becker, Portet, Pryzant, Qin,
  Radmilac, Ren, de~Rosa, Rosset, Roy, Ruwase, Saarikivi, Saied, Salim,
  Santacroce, Shah, Shang, Sharma, Shen, Shukla, Song, Tanaka, Tupini,
  Vaddamanu, Wang, Wang, Wang, Wang, Wang, Wang, Ward, Wen, Witte, Wu, Wu,
  Wyatt, Xiao, Xu, Xu, Xu, Xue, Yadav, Yang, Yang, Yang, Yang, Yu, Yuan, Zhang,
  Zhang, Zhang, Zhang, Zhang, Zhang, Zhang, and
  Zhou}]{Abdin-2024-Phi3-TechReport}
Marah Abdin, Jyoti Aneja, Hany Awadalla, Ahmed Awadallah, Ammar~Ahmad Awan,
  Nguyen Bach, Amit Bahree, Arash Bakhtiari, Jianmin Bao, Harkirat Behl, Alon
  Benhaim, Misha Bilenko, Johan Bjorck, Sébastien Bubeck, Martin Cai, Qin Cai,
  Vishrav Chaudhary, Dong Chen, Dongdong Chen, and 110 others. 2024.
\newblock \href {https://arxiv.org/abs/2404.14219} {Phi-3 technical report: A
  highly capable language model locally on your phone}.
\newblock \emph{Preprint}, arXiv:2404.14219.

\bibitem[{Brew(1992)}]{Brew-1992-Shake-and-Bake-MT}
Chris Brew. 1992.
\newblock \href {https://doi.org/10.3115/992133.992165} {Letting the cat out of
  the bag: generation for shake-and-bake mt}.
\newblock In \emph{Proceedings of the 14th Conference on Computational
  Linguistics - Volume 2}, COLING '92, page 610–616, USA. Association for
  Computational Linguistics.

\bibitem[{Brown et~al.(1990)Brown, Cocke, Della~Pietra, Della~Pietra, Jelinek,
  Lafferty, Mercer, and Roossin}]{BrownEtal-1990-SMT}
Peter~F. Brown, John Cocke, Stephen~A. Della~Pietra, Vincent~J. Della~Pietra,
  Fredrick Jelinek, John~D. Lafferty, Robert~L. Mercer, and Paul~S. Roossin.
  1990.
\newblock \href {https://aclanthology.org/J90-2002/} {A statistical approach to
  machine translation}.
\newblock \emph{Computational Linguistics}, 16(2):79--85.

\bibitem[{Elman(1990)}]{Elman-1990-FindingStructureInTime}
Jeffrey~L. Elman. 1990.
\newblock \href {https://doi.org/10.1016/0364-0213(90)90002-E} {Finding
  structure in time}.
\newblock \emph{Cognitive Science}, 14(2):179--211.

\bibitem[{{Gemma Team} et~al.(2026){Gemma Team}, Abd, Aggarwal, Algayres,
  Andreev, Bachem, Ballantyne, Brick, Cărbune, Casbon, Chaturvedi, Chawla,
  Cotruta, Coucke, Culliton, Dadashi, Dixon, Elhawaty, Evci, Farabet, Ferret,
  Galgani, Girgin, Grill, Grootendorst, Guo, Hardin, He, Hernandez, Homburger,
  Hussenot, Ji, Joulin, Kamath, Kassraie, Lacombe, Lahoti, Liu, Martins,
  Martins, Matejovicova, Merhej, Momchev, Mondal, Mullins, Panyam, Pathak,
  Perrin, Pinto, Pot, Pouget, Ramé, Ramos, Reid, Rim, Rivière, Roth,
  Rouillard, Sanseviero, Sessa, Settle, Sinopalnikov, Smoot, Stanczyk, Steiner,
  Stewart, Tolstikhin, Tschannen, Tsitsulin, Vieillard, Wu, Xu, Yang, Yvinec,
  Zhang, Zhang, Zou, Aagnes, Abdelhamed, Adamek, Agrawal, Agrawal,
  Alabdulmohsin, Alayrac, Alon, Amarnath, Anand, Anastasiou, Ariafar, Aubet,
  Axiotis, Barbero, Barral, Bendebury, Bergmann, Bileschi, Black, Blondel,
  Borgeaud, Bražinskas, Burnell, Busa-Fekete, Cai, Calandriello, Cameron,
  Caucheteux, Chaabouni, Chadha, Chan, Chen, Chen, Chen, Chen, Cheng, hsiang
  Chien, Chinaev, Chou, Chu, Coleman, Consul, Conway-Rahman, Crowell, Cutler,
  Dani, Daruki, Das, Deutsch, Dikkala, Ding, Ding, Dodhia, Donhauser, Doshi,
  Dragan, Druinsky, Dua, Egyed, Eisenbud, Eppens, Fan, Fatemi, Fathullah,
  Feinberg, Ferev, Flennerhag, Fujimoto, Oliveira, Galatzer-Levy, Gante,
  Geisler, Ghosal, Girgis, von Glehn, Go, Gokhale, Grills, Gu, Gupta, Gupta,
  Guruganesh, Hadsell, Harkous, Harlalka, Hassabis, Hauth, Heyward, Hosseini,
  Hsia, Hsu, Huang, Huang, Hui, Hutter, I, Iliopoulos, Jain, Jawahar, Ji, Jin,
  Johnson, Joshi, Kandoor, Kang, Kavukcuoglu, Kazemi, Kenealy, Khalifa, Kirk,
  Korotkov, Kothawade, Kovalev, Kovelamudi, Kraft, Kumar, Kumar, Kuppam,
  Lannin, Lee, Lee, Lepikhin, Levkovitch, Li, Li, Liévin, Lin, Lin, Liu, Liu,
  Liu, Liu, Lobov, Lunayach, Ma, Madan, Maksai, Malmi, Matuszak, McDuff,
  Menghani, Mikuła, Mirylenka, Misiunas, Misra, Mitran, Mohamed, Mukha,
  Noland, O'Donnell, O'Donoghue, Olszewska, Orlando, Pan, Panigrahy, Parekh,
  Perez-Nieves, Park, Paskie, Peng, Petrini, Petrov, Pfeiffer, Piot, Plomecka,
  Poder, Ponce, Pramanik, Racz, Rajan, Ramanovich, Rao, Ritter, Rodrigues,
  Rosen, Rybiński, Sachdeva, Sander, Sathyanarayana, Savla, Schmidgall,
  Schuster, Scrivener, Seguin, Sellergren, Severyn, Shafran, Shah, Shahriari,
  Shangguan, Shenoy, Shenoy, Shivanna, Sho, Spangher, Stokowiec, Strother, Su,
  Sun, Sundararajan, Tacchetti, Taege, Tafti, Tarbouriech, Tekur, Thakoor,
  Thapa, Traverse, Treven, Tu, Tung, Çağlar Ünlü, Veličković, Venkat,
  Venkatesh, Venkiteswaran, Visin, Vitvitskyi, Vodrahalli, Wang, Wang,
  Warkentin, Wassenberg, Wieting, Wu, Xiao, Xu, Xu, Xue, Yadav, Yan, Yang,
  Yang, Yang, Ying, Yoo, Zadimoghaddam, Zafar, Zhang, Zhang, Zhang, Zhang,
  Zhao, Zhou, and Zou}]{Google-GemmaTeam-2026-Gemma4-TechReport}
{Gemma Team}, Sherif~El Abd, Vaibhav Aggarwal, Robin Algayres, Alek Andreev,
  Olivier Bachem, Ian Ballantyne, Cormac Brick, Victor Cărbune, Michelle
  Casbon, Mayank Chaturvedi, Aditya Chawla, Victor Cotruta, Alice Coucke, Phil
  Culliton, Robert Dadashi, Lucas Dixon, Mohamed Elhawaty, Utku Evci, and 304
  others. 2026.
\newblock \href {https://arxiv.org/abs/2607.02770} {Gemma 4 technical report}.
\newblock \emph{Preprint}, arXiv:2607.02770.

\bibitem[{Hu et~al.(2022)Hu, Shen, Wallis, Allen-Zhu, Li, Wang, Wang, and
  Chen}]{HuEtAl-2022-LoRA-Fine-tuning}
Edward~J Hu, Yelong Shen, Phillip Wallis, Zeyuan Allen-Zhu, Yuanzhi Li, Shean
  Wang, Lu~Wang, and Weizhu Chen. 2022.
\newblock \href {https://openreview.net/forum?id=nZeVKeeFYf9} {Lo{RA}: Low-rank
  adaptation of large language models}.
\newblock In \emph{International Conference on Learning Representations}.

\bibitem[{IEEE(1969)}]{IEEE-Classic-Harvard-Speach-Quality-Sentences-1969}
IEEE. 1969.
\newblock \href {https://doi.org/10.1109/IEEESTD.1969.7405210} {Ieee
  recommended practice for speech quality measurements}.
\newblock \emph{IEEE No 297-1969}, pages 1--24.

\bibitem[{Kingma and Ba(2015)}]{Kingma-2017-Adam-Optimizer}
Diederik~P. Kingma and Jimmy Ba. 2015.
\newblock \href {http://arxiv.org/abs/1412.6980} {Adam: {A} method for
  stochastic optimization}.
\newblock In \emph{3rd International Conference on Learning Representations,
  {ICLR} 2015, San Diego, CA, USA, May 7-9, 2015, Conference Track
  Proceedings}.

\bibitem[{Loshchilov and Hutter(2019)}]{LoshchilovAndHutter-2019-AdamW}
Ilya Loshchilov and Frank Hutter. 2019.
\newblock \href {https://openreview.net/forum?id=Bkg6RiCqY7} {Decoupled weight
  decay regularization}.
\newblock In \emph{International Conference on Learning Representations}.

\bibitem[{Ouyang et~al.(2022)Ouyang, Wu, Jiang, Almeida, Wainwright, Mishkin,
  Zhang, Agarwal, Slama, Ray, Schulman, Hilton, Kelton, Miller, Simens, Askell,
  Welinder, Christiano, Leike, and Lowe}]{OuyangEtAl-2022-RLHF}
Long Ouyang, Jeffrey Wu, Xu~Jiang, Diogo Almeida, Carroll Wainwright, Pamela
  Mishkin, Chong Zhang, Sandhini Agarwal, Katarina Slama, Alex Ray, John
  Schulman, Jacob Hilton, Fraser Kelton, Luke Miller, Maddie Simens, Amanda
  Askell, Peter Welinder, Paul~F Christiano, Jan Leike, and Ryan Lowe. 2022.
\newblock \href
  {https://proceedings.neurips.cc/paper_files/paper/2022/file/b1efde53be364a73914f58805a001731-Paper-Conference.pdf}
  {Training language models to follow instructions with human feedback}.
\newblock In \emph{Advances in Neural Information Processing Systems},
  volume~35, pages 27730--27744. Curran Associates, Inc.

\bibitem[{Rafailov et~al.(2023)Rafailov, Sharma, Mitchell, Manning, Ermon, and
  Finn}]{Rafailov-EtAl-2023-DPO}
Rafael Rafailov, Archit Sharma, Eric Mitchell, Christopher~D Manning, Stefano
  Ermon, and Chelsea Finn. 2023.
\newblock \href
  {https://proceedings.neurips.cc/paper_files/paper/2023/file/a85b405ed65c6477a4fe8302b5e06ce7-Paper-Conference.pdf}
  {Direct preference optimization: Your language model is secretly a reward
  model}.
\newblock In \emph{Advances in Neural Information Processing Systems},
  volume~36, pages 53728--53741. Curran Associates, Inc.

\bibitem[{Reimers and
  Gurevych(2019)}]{ReimersAndGurevych-2019-SentenceTransformers}
Nils Reimers and Iryna Gurevych. 2019.
\newblock \href {https://doi.org/10.18653/v1/D19-1410} {Sentence-{BERT}:
  Sentence embeddings using {S}iamese {BERT}-networks}.
\newblock In \emph{Proceedings of the 2019 Conference on Empirical Methods in
  Natural Language Processing and the 9th International Joint Conference on
  Natural Language Processing (EMNLP-IJCNLP)}, pages 3982--3992, Hong Kong,
  China. Association for Computational Linguistics.

\bibitem[{Schmaltz(2026{\natexlab{a}})}]{Schmaltz-2026-ReexpressMCPServer}
Allen Schmaltz. 2026{\natexlab{a}}.
\newblock \href {https://doi.org/10.1145/3786335.3813214} {Introspectable,
  updatable, and uncertainty-aware classification of language model
  instruction-following}.
\newblock In \emph{Proceedings of the ACM Conference on AI and Agentic
  Systems}, CAIS '26, page 1259–1269, New York, NY, USA. Association for
  Computing Machinery.

\bibitem[{Schmaltz(2026{\natexlab{b}})}]{Schmaltz-2026-SDM-Activations}
Allen Schmaltz. 2026{\natexlab{b}}.
\newblock \href {https://doi.org/10.18653/v1/2026.findings-acl.1109}
  {Similarity-distance-magnitude activations}.
\newblock In \emph{Findings of the {A}ssociation for {C}omputational
  {L}inguistics: {ACL} 2026}, pages 22037--22057, San Diego, California, United
  States. Association for Computational Linguistics.

\bibitem[{Schmaltz et~al.(2016{\natexlab{a}})Schmaltz, Kim, Rush, and
  Shieber}]{Schmaltz-Etal-2016-ClassificationAsTransduction}
Allen Schmaltz, Yoon Kim, Alexander~M. Rush, and Stuart Shieber.
  2016{\natexlab{a}}.
\newblock \href {https://doi.org/10.18653/v1/W16-0528} {Sentence-level
  grammatical error identification as sequence-to-sequence correction}.
\newblock In \emph{Proceedings of the 11th Workshop on Innovative Use of {NLP}
  for Building Educational Applications}, pages 242--251, San Diego, CA.
  Association for Computational Linguistics.

\bibitem[{Schmaltz et~al.(2016{\natexlab{b}})Schmaltz, Rush, and
  Shieber}]{Schmaltz-EtAl-2016-WordOrderingWithoutSyntax}
Allen Schmaltz, Alexander~M. Rush, and Stuart Shieber. 2016{\natexlab{b}}.
\newblock \href {https://doi.org/10.18653/v1/D16-1255} {Word ordering without
  syntax}.
\newblock In \emph{Proceedings of the 2016 Conference on Empirical Methods in
  Natural Language Processing}, pages 2319--2324, Austin, Texas. Association
  for Computational Linguistics.

\bibitem[{Schulman et~al.(2017)Schulman, Wolski, Dhariwal, Radford, and
  Klimov}]{SchulmanEtAl-2017-PPO}
John Schulman, Filip Wolski, Prafulla Dhariwal, Alec Radford, and Oleg Klimov.
  2017.
\newblock \href {https://arxiv.org/abs/1707.06347} {Proximal policy
  optimization algorithms}.
\newblock \emph{Preprint}, arXiv:1707.06347.

\bibitem[{Vaswani et~al.(2017)Vaswani, Shazeer, Parmar, Uszkoreit, Jones,
  Gomez, Kaiser, and Polosukhin}]{VaswaniEtAl-2017-AttentionIsAllYouNeed}
Ashish Vaswani, Noam Shazeer, Niki Parmar, Jakob Uszkoreit, Llion Jones,
  Aidan~N. Gomez, \L{}ukasz Kaiser, and Illia Polosukhin. 2017.
\newblock Attention is all you need.
\newblock In \emph{Proceedings of the 31st International Conference on Neural
  Information Processing Systems}, NIPS'17, page 6000–6010, Red Hook, NY,
  USA. Curran Associates Inc.

\bibitem[{Zhang et~al.(2025)Zhang, Hosseini, Bansal, Kazemi, Kumar, and
  Agarwal}]{ZhangEtlAl-2025-Generative}
Lunjun Zhang, Arian Hosseini, Hritik Bansal, Mehran Kazemi, Aviral Kumar, and
  Rishabh Agarwal. 2025.
\newblock \href {https://openreview.net/forum?id=Ccwp4tFEtE} {Generative
  verifiers: Reward modeling as next-token prediction}.
\newblock In \emph{The Thirteenth International Conference on Learning
  Representations}.

\end{thebibliography}


\appendix

\section{Appendix}
\label{sec:appendix}

We provide additional training details in Appendix~\ref{appendix:additional-implementation-details}; Appendix~\ref{sec:positive-only-prompt} provides the prompt used for the models trained with positive-only sequences; and Appendix~\ref{sec:full-conditional-accuracy-tables} provides additional results stratified by the class- and prediction-conditional accuracy.

\begin{table*}
  \centering
  \setlength{\fboxsep}{1pt}
  \resizebox{0.82\textwidth}{!}{%
  \begin{tabular}{l l  l c c c c c c c c c l }
    \toprule

    & & &  \multicolumn{4}{c}{Class-conditional} & \multicolumn{4}{c}{Prediction-conditional}  & \multicolumn{2}{c}{Marginal} \\
    & & &  \multicolumn{2}{c}{$y=0$} & \multicolumn{2}{c}{$y=1$} & \multicolumn{2}{c}{$\hat{y}=0$} & \multicolumn{2}{c}{$\hat{y}=1$} & \multicolumn{2}{c}{$y\in \{0,1\}$} \\
    \cmidrule(r){4-5} \cmidrule(r){6-7} \cmidrule(r){8-9} \cmidrule(r){10-11} \cmidrule(r){12-13}
    Dataset   & Model & Estimator & \textsc{Acc.}& $\frac{n}{|\testSplit|}$ & \textsc{Acc.} & $\frac{n}{|\testSplit|}$ & \textsc{Acc.} & $\frac{n}{|\testSplit|}$ & \textsc{Acc.} & $\frac{n}{|\testSplit|}$ & \textsc{Acc.} & $\frac{n}{|\testSplit|}$\\

\midrule
$\datasetLMOrderHarvard$ & $\modelPhiThreeFiveInstruct$ & $\sdmNested$ & \colorbox{correctPredictionColor}{0.96} & 0.14 & \colorbox{correctPredictionColor}{1.0} & 0.042 & \colorbox{correctPredictionColor}{1.0} & 0.14 & \colorbox{wrongPredictionColor}{0.86} & 0.048 & \colorbox{correctPredictionColor}{0.97} & 0.19 \\
$\datasetLMOrderHarvard$ & $\modelPhiThreeFiveInstructFinetuned$ & $\sdmNested$ & \colorbox{correctPredictionColor}{\allRejected} & 0.0 & \colorbox{correctPredictionColor}{\allRejected} & 0.0 & \colorbox{correctPredictionColor}{\allRejected} & 0.0 & \colorbox{correctPredictionColor}{\allRejected} & 0.0 & \colorbox{correctPredictionColor}{\allRejected} & 0.0 \\
$\datasetLMOrderHarvard$ & $\modelPhiThreeFiveInstructFinetunedAdditionalRun$ & $\sdmNested$ & \colorbox{correctPredictionColor}{\allRejected} & 0.0 & \colorbox{correctPredictionColor}{\allRejected} & 0.0 & \colorbox{correctPredictionColor}{\allRejected} & 0.0 & \colorbox{correctPredictionColor}{\allRejected} & 0.0 & \colorbox{correctPredictionColor}{\allRejected} & 0.0 \\
$\datasetLMOrderHarvard$ & $\modelPhiThreeFiveInstructFinetunedCM$ & $\sdmNested$ & \colorbox{wrongPredictionColor}{0.0} & 0.0025 & \colorbox{correctPredictionColor}{1.0} & 0.11 & \colorbox{correctPredictionColor}{\allRejected} & 0.0 & \colorbox{wrongPredictionColor}{0.95} & 0.11 & \colorbox{wrongPredictionColor}{0.95} & 0.11 \\
$\datasetLMOrderHarvard$ & $\modelPhiThreeFiveInstructFinetunedCMAdditionalRun$ & $\sdmNested$ & \colorbox{wrongPredictionColor}{0.0} & 0.0012 & \colorbox{correctPredictionColor}{1.0} & 0.27 & \colorbox{correctPredictionColor}{\allRejected} & 0.0 & \colorbox{correctPredictionColor}{1.0} & 0.27 & \colorbox{correctPredictionColor}{1.0} & 0.27 \\

\midrule
$\datasetLMOrderHarvard$ & $\modelGemmaFourEtwoBInstruct$ & $\sdmNested$ & \colorbox{correctPredictionColor}{1.0} & 0.87 & \colorbox{correctPredictionColor}{\allRejected} & 0.0 & \colorbox{correctPredictionColor}{1.0} & 0.87 & \colorbox{wrongPredictionColor}{0.0} & 0.00069 & \colorbox{correctPredictionColor}{1.0} & 0.87 \\
$\datasetLMOrderHarvard$ & $\modelGemmaFourEtwoBInstructFinetuned$ & $\sdmNested$ & \colorbox{correctPredictionColor}{\allRejected} & 0.0 & \colorbox{correctPredictionColor}{1.0} & 0.064 & \colorbox{correctPredictionColor}{\allRejected} & 0.0 & \colorbox{correctPredictionColor}{1.0} & 0.064 & \colorbox{correctPredictionColor}{1.0} & 0.064 \\
$\datasetLMOrderHarvard$ & $\modelGemmaFourEtwoBInstructFinetunedAdditionalRun$ & $\sdmNested$ & \colorbox{correctPredictionColor}{1.0} & 0.0014 & \colorbox{correctPredictionColor}{1.0} & 0.0014 & \colorbox{correctPredictionColor}{1.0} & 0.0014 & \colorbox{correctPredictionColor}{1.0} & 0.0014 & \colorbox{correctPredictionColor}{1.0} & 0.0028 \\
$\datasetLMOrderHarvard$ & $\modelGemmaFourEtwoBInstructFinetunedCM$ & $\sdmNested$ & \colorbox{correctPredictionColor}{1.0} & 0.0014 & \colorbox{correctPredictionColor}{1.0} & 0.036 & \colorbox{correctPredictionColor}{1.0} & 0.0014 & \colorbox{correctPredictionColor}{1.0} & 0.036 & \colorbox{correctPredictionColor}{1.0} & 0.038 \\
$\datasetLMOrderHarvard$ & $\modelGemmaFourEtwoBInstructFinetunedCMAdditionalRun$ & $\sdmNested$ & \colorbox{correctPredictionColor}{1.0} & 0.00023 & \colorbox{correctPredictionColor}{1.0} & 0.10 & \colorbox{correctPredictionColor}{1.0} & 0.00023 & \colorbox{correctPredictionColor}{1.0} & 0.10 & \colorbox{correctPredictionColor}{1.0} & 0.10 \\

\midrule
$\datasetLMOrderWikipedia$ & $\modelPhiThreeFiveInstruct$ & $\sdmNested$ & \colorbox{correctPredictionColor}{1.0} & 0.13 & \colorbox{correctPredictionColor}{1.0} & 0.0027 & \colorbox{correctPredictionColor}{1.0} & 0.13 & \colorbox{wrongPredictionColor}{0.90} & 0.0031 & \colorbox{correctPredictionColor}{1.0} & 0.13 \\
$\datasetLMOrderWikipedia$ & $\modelPhiThreeFiveInstructFinetuned$ & $\sdmNested$ & \colorbox{correctPredictionColor}{1.0} & 0.0010 & \colorbox{correctPredictionColor}{1.0} & 0.0020 & \colorbox{correctPredictionColor}{1.0} & 0.0010 & \colorbox{correctPredictionColor}{1.0} & 0.0020 & \colorbox{correctPredictionColor}{1.0} & 0.0030 \\
$\datasetLMOrderWikipedia$ & $\modelPhiThreeFiveInstructFinetunedAdditionalRun$ & $\sdmNested$ & \colorbox{correctPredictionColor}{1.0} & 0.0018 & \colorbox{correctPredictionColor}{1.0} & 0.00067 & \colorbox{correctPredictionColor}{1.0} & 0.0018 & \colorbox{correctPredictionColor}{1.0} & 0.00067 & \colorbox{correctPredictionColor}{1.0} & 0.0024 \\
$\datasetLMOrderWikipedia$ & $\modelPhiThreeFiveInstructFinetunedCM$ & $\sdmNested$ & \colorbox{wrongPredictionColor}{0.72} & 0.0015 & \colorbox{correctPredictionColor}{1.0} & 0.093 & \colorbox{correctPredictionColor}{1.0} & 0.0010 & \colorbox{correctPredictionColor}{1.0} & 0.094 & \colorbox{correctPredictionColor}{1.0} & 0.095 \\
$\datasetLMOrderWikipedia$ & $\modelPhiThreeFiveInstructFinetunedCMAdditionalRun$ & $\sdmNested$ & \colorbox{wrongPredictionColor}{0.91} & 0.0027 & \colorbox{correctPredictionColor}{1.0} & 0.082 & \colorbox{correctPredictionColor}{1.0} & 0.0024 & \colorbox{correctPredictionColor}{1.0} & 0.082 & \colorbox{correctPredictionColor}{1.0} & 0.084 \\

\midrule
$\datasetLMOrderWikipedia$ & $\modelGemmaFourEtwoBInstruct$ & $\sdmNested$ & \colorbox{correctPredictionColor}{1.0} & 0.78 & \colorbox{correctPredictionColor}{1.0} & 0.0010 & \colorbox{correctPredictionColor}{1.0} & 0.78 & \colorbox{wrongPredictionColor}{0.32} & 0.0032 & \colorbox{correctPredictionColor}{1.0} & 0.78 \\
$\datasetLMOrderWikipedia$ & $\modelGemmaFourEtwoBInstructFinetuned$ & $\sdmNested$ & \colorbox{wrongPredictionColor}{0.83} & 0.00067 & \colorbox{correctPredictionColor}{1.0} & 0.036 & \colorbox{correctPredictionColor}{1.0} & 0.00050 & \colorbox{correctPredictionColor}{1.0} & 0.036 & \colorbox{correctPredictionColor}{1.0} & 0.037 \\
$\datasetLMOrderWikipedia$ & $\modelGemmaFourEtwoBInstructFinetunedAdditionalRun$ & $\sdmNested$ & \colorbox{correctPredictionColor}{1.0} & 0.0010 & \colorbox{correctPredictionColor}{1.0} & 0.0048 & \colorbox{correctPredictionColor}{1.0} & 0.0010 & \colorbox{correctPredictionColor}{1.0} & 0.0048 & \colorbox{correctPredictionColor}{1.0} & 0.0058 \\
$\datasetLMOrderWikipedia$ & $\modelGemmaFourEtwoBInstructFinetunedCM$ & $\sdmNested$ & \colorbox{wrongPredictionColor}{0.92} & 0.00058 & \colorbox{correctPredictionColor}{0.99} & 0.050 & \colorbox{wrongPredictionColor}{0.67} & 0.00083 & \colorbox{correctPredictionColor}{1.0} & 0.049 & \colorbox{correctPredictionColor}{0.99} & 0.050 \\
$\datasetLMOrderWikipedia$ & $\modelGemmaFourEtwoBInstructFinetunedCMAdditionalRun$ & $\sdmNested$ & \colorbox{correctPredictionColor}{1.0} & 0.00092 & \colorbox{correctPredictionColor}{1.0} & 0.035 & \colorbox{correctPredictionColor}{1.0} & 0.00092 & \colorbox{correctPredictionColor}{1.0} & 0.035 & \colorbox{correctPredictionColor}{1.0} & 0.036 \\

\midrule
$\datasetLMOrderWikipediaCVS$ & $\modelPhiThreeFiveInstruct$ & $\sdmNested$ & \colorbox{correctPredictionColor}{1.0} & 0.020 & \colorbox{correctPredictionColor}{\allRejected} & 0.0 & \colorbox{correctPredictionColor}{1.0} & 0.020 & \colorbox{correctPredictionColor}{\allRejected} & 0.0 & \colorbox{correctPredictionColor}{1.0} & 0.020 \\
$\datasetLMOrderWikipediaCVS$ & $\modelPhiThreeFiveInstructFinetuned$ & $\sdmNested$ & \colorbox{correctPredictionColor}{\allRejected} & 0.0 & \colorbox{correctPredictionColor}{\allRejected} & 0.0 & \colorbox{correctPredictionColor}{\allRejected} & 0.0 & \colorbox{correctPredictionColor}{\allRejected} & 0.0 & \colorbox{correctPredictionColor}{\allRejected} & 0.0 \\
$\datasetLMOrderWikipediaCVS$ & $\modelPhiThreeFiveInstructFinetunedAdditionalRun$ & $\sdmNested$ & \colorbox{correctPredictionColor}{\allRejected} & 0.0 & \colorbox{correctPredictionColor}{\allRejected} & 0.0 & \colorbox{correctPredictionColor}{\allRejected} & 0.0 & \colorbox{correctPredictionColor}{\allRejected} & 0.0 & \colorbox{correctPredictionColor}{\allRejected} & 0.0 \\
$\datasetLMOrderWikipediaCVS$ & $\modelPhiThreeFiveInstructFinetunedCM$ & $\sdmNested$ & \colorbox{correctPredictionColor}{1.0} & 0.00042 & \colorbox{correctPredictionColor}{\allRejected} & 0.0 & \colorbox{correctPredictionColor}{1.0} & 0.00042 & \colorbox{correctPredictionColor}{\allRejected} & 0.0 & \colorbox{correctPredictionColor}{1.0} & 0.00042 \\
$\datasetLMOrderWikipediaCVS$ & $\modelPhiThreeFiveInstructFinetunedCMAdditionalRun$ & $\sdmNested$ & \colorbox{correctPredictionColor}{1.0} & 0.00050 & \colorbox{correctPredictionColor}{\allRejected} & 0.0 & \colorbox{correctPredictionColor}{1.0} & 0.00050 & \colorbox{correctPredictionColor}{\allRejected} & 0.0 & \colorbox{correctPredictionColor}{1.0} & 0.00050 \\

\midrule
$\datasetLMOrderWikipediaCVS$ & $\modelGemmaFourEtwoBInstruct$ & $\sdmNested$ & \colorbox{correctPredictionColor}{0.99} & 0.33 & \colorbox{correctPredictionColor}{\allRejected} & 0.0 & \colorbox{correctPredictionColor}{1.0} & 0.33 & \colorbox{wrongPredictionColor}{0.0} & 0.0026 & \colorbox{correctPredictionColor}{0.99} & 0.33 \\
$\datasetLMOrderWikipediaCVS$ & $\modelGemmaFourEtwoBInstructFinetuned$ & $\sdmNested$ & \colorbox{correctPredictionColor}{\allRejected} & 0.0 & \colorbox{correctPredictionColor}{\allRejected} & 0.0 & \colorbox{correctPredictionColor}{\allRejected} & 0.0 & \colorbox{correctPredictionColor}{\allRejected} & 0.0 & \colorbox{correctPredictionColor}{\allRejected} & 0.0 \\
$\datasetLMOrderWikipediaCVS$ & $\modelGemmaFourEtwoBInstructFinetunedAdditionalRun$ & $\sdmNested$ & \colorbox{correctPredictionColor}{\allRejected} & 0.0 & \colorbox{correctPredictionColor}{\allRejected} & 0.0 & \colorbox{correctPredictionColor}{\allRejected} & 0.0 & \colorbox{correctPredictionColor}{\allRejected} & 0.0 & \colorbox{correctPredictionColor}{\allRejected} & 0.0 \\
$\datasetLMOrderWikipediaCVS$ & $\modelGemmaFourEtwoBInstructFinetunedCM$ & $\sdmNested$ & \colorbox{correctPredictionColor}{\allRejected} & 0.0 & \colorbox{correctPredictionColor}{1.0} & 0.0025 & \colorbox{correctPredictionColor}{\allRejected} & 0.0 & \colorbox{correctPredictionColor}{1.0} & 0.0025 & \colorbox{correctPredictionColor}{1.0} & 0.0025 \\
$\datasetLMOrderWikipediaCVS$ & $\modelGemmaFourEtwoBInstructFinetunedCMAdditionalRun$ & $\sdmNested$ & \colorbox{correctPredictionColor}{\allRejected} & 0.0 & \colorbox{correctPredictionColor}{1.0} & 0.00017 & \colorbox{correctPredictionColor}{\allRejected} & 0.0 & \colorbox{correctPredictionColor}{1.0} & 0.00017 & \colorbox{correctPredictionColor}{1.0} & 0.00017 \\

    \bottomrule
  \end{tabular}
  }  
    \caption{Selective classification results for instruction-following over generations using nested $\sdm$ estimators with resolution $r=0.01$. The admitted set includes every available region with $\alpha\ge0.95$, even when no region exists at exactly $0.95$. Results are averaged over six independently calibrated test-time $\sdm$ estimators, with LM seeds kept separate. The corresponding $\estimatorNoReject$ rows are omitted, because they are the same as those in Table~\ref{tab:experiments-verification-results}.} %
  \label{tab:experiments-verification-results-resolution01nested95} 
\end{table*}

\subsection{Additional Implementation Details}\label{appendix:additional-implementation-details}

In all experiments, we fine-tune using LoRA adaptors \cite{HuEtAl-2022-LoRA-Fine-tuning}, with rank 16 and a scaling parameter of 16, applied to the attention and feed-forward projections in the Transformer \cite{VaswaniEtAl-2017-AttentionIsAllYouNeed}. Dropout is not used, and the LM's final linear layer over $\gV$ remains frozen. For updating the parameters associated with the LoRA adaptors (i.e., excluding the $\sdm$ layers), we use the AdamW optimizer \citep{LoshchilovAndHutter-2019-AdamW} with a constant learning rate of $1 \times 10^{-4}$ for training and no weight decay for a total of 10 epochs. Each optimizer update contains two underlying problems, which corresponds to two response sequences for the \textsc{+FT} models or four sequences for the \textsc{+CM} models, where only the final verification tag sequences are supervised for the negative sequences. In this way, each epoch across experimental settings consists of 2,500 optimizer updates, and positive-sequence exposure is matched; however, CM also processes negative sequences. We repeat fine-tuning twice for each experimental setting, using separate seeds. Aside from the input encoding of the \textsc{+CM} models, these fine-tuning settings are intended to be typical LoRA settings for supervised fine-tuning at this scale.

We use version 0.4.9 of the \texttt{reexpress\_sdm} Python package\footnote{\url{https://pypi.org/project/reexpress-sdm}} with exemplar dimension 1000 and \texttt{max\_neighbors=2048}. Inputs are standardized using the mean and standard deviation over the $\sdm$ activation's training set input vectors. Concatenating the mean-pooled and final-token hidden states gives 6,144 input dimensions for \texttt{Phi-3.5-mini-instruct} and 3,072 for \texttt{Gemma-4-E2B-it}. For \textsc{+FT} models, the final-token boundary is taken as the end of \texttt{</sentence>}, including when teacher-forcing static negatives. The other models use the \texttt{<verified>} boundary described in the main text.

All $\sdm$ activations are trained and calibrated by holding the parameters of the underlying LM fixed. Each training iteration uses 500 epochs, with standard cross-entropy in the first epoch and the $\sdm$ loss thereafter, using the Adam optimizer \citep{Kingma-2017-Adam-Optimizer} with learning rate $10^{-5}$, batch size 64, and no weight decay. We chose the epochs with the lowest class-balanced $\sdm$ loss over the calibration set. The $\sdm$ estimators for LM epoch selection use one fixed initialization, $J=1$. The test-time estimators are instead trained and calibrated over six independent initializations with fixed training/calibration sets. Accuracy averages in the tables omit estimators with an empty corresponding $\alpha$ region, whereas admission averages include all six estimators, counting empty regions as zero. The $\sdm$ activations themselves have 3-6 million parameters, depending on the LM (and by extension, the hidden state dimensions).

Each model's LoRA fine-tuning and generations used one A100 GPU with 40 GB of memory. Some of the test-time $\sdm$ estimators were trained on an M2 Ultra Mac with 128 GB of memory.

\subsection{Prompts}\label{sec:positive-only-prompt}

The prompt for the base models and the models fine-tuned with $\encodingContrastiveMasking$ appear in Table~\ref{tab:contrastive-encoding}, substituting the sentence prefix and shuffled suffix for each problem. The models fine-tuned only with positive sequences omit the final verification tag instruction, as follows:
\begin{quote}
Complete the sentence `Neat plans' by reordering all of the following without adding new punctuation nor words: `fail luck. without'. Only reply with the sentence in the XML <sentence> </sentence>.
\end{quote}
\subsection{Additional Reference Results}\label{sec:full-conditional-accuracy-tables}

For reference, Table~\ref{tab:experiments-verification-results} is an expanded version of Table~\ref{tab:experiments-verification-results-abbreviated} including the class- and prediction-conditional results. Table~\ref{tab:experiments-verification-results-resolution01nested95} uses the same $\sdm$ estimator as Table~\ref{tab:experiments-verification-results}, but with the results using the less stringent filtering of any nested region with $\alpha \ge 0.95$. Finally, Table~\ref{tab:experiments-verification-results-static} is the expanded version of Table~\ref{tab:experiments-verification-results-abbreviated-static-resolution01-hr99}. 

\begin{table*}
  \centering
  \setlength{\fboxsep}{1pt}
  \resizebox{0.82\textwidth}{!}{%
  \begin{tabular}{l l  l c c c c c c c c c l }
    \toprule

    & & &  \multicolumn{4}{c}{Class-conditional} & \multicolumn{4}{c}{Prediction-conditional}  & \multicolumn{2}{c}{Marginal} \\
    & & &  \multicolumn{2}{c}{$y=0$} & \multicolumn{2}{c}{$y=1$} & \multicolumn{2}{c}{$\hat{y}=0$} & \multicolumn{2}{c}{$\hat{y}=1$} & \multicolumn{2}{c}{$y\in \{0,1\}$} \\
    \cmidrule(r){4-5} \cmidrule(r){6-7} \cmidrule(r){8-9} \cmidrule(r){10-11} \cmidrule(r){12-13}
    Dataset   & Model & Estimator & \textsc{Acc.}& $\frac{n}{|\testSplit|}$ & \textsc{Acc.} & $\frac{n}{|\testSplit|}$ & \textsc{Acc.} & $\frac{n}{|\testSplit|}$ & \textsc{Acc.} & $\frac{n}{|\testSplit|}$ & \textsc{Acc.} & $\frac{n}{|\testSplit|}$\\
   
   \midrule
  
$\datasetLMOrderHarvard$ & $\modelPhiThreeFiveInstruct$ & $\estimatorNoReject$ & \colorbox{wrongPredictionColor}{0.77} & 0.55 & \colorbox{wrongPredictionColor}{0.95} & 0.42 & \colorbox{wrongPredictionColor}{0.96} & 0.45 & \colorbox{wrongPredictionColor}{0.76} & 0.53 & \colorbox{wrongPredictionColor}{0.85} & 0.98 \\
$\datasetLMOrderHarvard$ & $\modelPhiThreeFiveInstructFinetuned$ & $\estimatorNoReject$ & \colorbox{wrongPredictionColor}{0.53} & 0.063 & \colorbox{wrongPredictionColor}{0.97} & 0.94 & \colorbox{wrongPredictionColor}{0.56} & 0.059 & \colorbox{wrongPredictionColor}{0.97} & 0.94 & \colorbox{wrongPredictionColor}{0.94} & 1.0 \\
$\datasetLMOrderHarvard$ & $\modelPhiThreeFiveInstructFinetunedAdditionalRun$ & $\estimatorNoReject$ & \colorbox{wrongPredictionColor}{0.41} & 0.064 & \colorbox{wrongPredictionColor}{0.96} & 0.94 & \colorbox{wrongPredictionColor}{0.40} & 0.066 & \colorbox{wrongPredictionColor}{0.96} & 0.93 & \colorbox{wrongPredictionColor}{0.92} & 1.0 \\
$\datasetLMOrderHarvard$ & $\modelPhiThreeFiveInstructFinetunedCM$ & $\estimatorNoReject$ & \colorbox{wrongPredictionColor}{0.73} & 0.064 & \colorbox{wrongPredictionColor}{0.94} & 0.93 & \colorbox{wrongPredictionColor}{0.45} & 0.10 & \colorbox{wrongPredictionColor}{0.98} & 0.89 & \colorbox{wrongPredictionColor}{0.92} & 1.0 \\
$\datasetLMOrderHarvard$ & $\modelPhiThreeFiveInstructFinetunedCMAdditionalRun$ & $\estimatorNoReject$ & \colorbox{wrongPredictionColor}{0.65} & 0.065 & \colorbox{wrongPredictionColor}{0.98} & 0.93 & \colorbox{wrongPredictionColor}{0.68} & 0.063 & \colorbox{wrongPredictionColor}{0.98} & 0.94 & \colorbox{wrongPredictionColor}{0.96} & 1.0 \\

\midrule
$\datasetLMOrderHarvard$ & $\modelGemmaFourEtwoBInstruct$ & $\estimatorNoReject$ & \colorbox{correctPredictionColor}{0.99} & 1.0 & \textemdash & 0.0 & \colorbox{correctPredictionColor}{1.0} & 0.99 & \colorbox{wrongPredictionColor}{0.0} & 0.0058 & \colorbox{correctPredictionColor}{0.99} & 1.0 \\
$\datasetLMOrderHarvard$ & $\modelGemmaFourEtwoBInstructFinetuned$ & $\estimatorNoReject$ & \colorbox{wrongPredictionColor}{0.51} & 0.060 & \colorbox{wrongPredictionColor}{0.95} & 0.94 & \colorbox{wrongPredictionColor}{0.41} & 0.073 & \colorbox{wrongPredictionColor}{0.97} & 0.93 & \colorbox{wrongPredictionColor}{0.93} & 1.0 \\
$\datasetLMOrderHarvard$ & $\modelGemmaFourEtwoBInstructFinetunedAdditionalRun$ & $\estimatorNoReject$ & \colorbox{wrongPredictionColor}{0.76} & 0.057 & \colorbox{wrongPredictionColor}{0.90} & 0.94 & \colorbox{wrongPredictionColor}{0.31} & 0.14 & \colorbox{wrongPredictionColor}{0.98} & 0.86 & \colorbox{wrongPredictionColor}{0.89} & 1.0 \\
$\datasetLMOrderHarvard$ & $\modelGemmaFourEtwoBInstructFinetunedCM$ & $\estimatorNoReject$ & \colorbox{wrongPredictionColor}{0.67} & 0.040 & \colorbox{wrongPredictionColor}{0.95} & 0.96 & \colorbox{wrongPredictionColor}{0.37} & 0.074 & \colorbox{wrongPredictionColor}{0.99} & 0.93 & \colorbox{wrongPredictionColor}{0.94} & 1.0 \\
$\datasetLMOrderHarvard$ & $\modelGemmaFourEtwoBInstructFinetunedCMAdditionalRun$ & $\estimatorNoReject$ & \colorbox{wrongPredictionColor}{0.56} & 0.060 & \colorbox{wrongPredictionColor}{0.97} & 0.94 & \colorbox{wrongPredictionColor}{0.56} & 0.060 & \colorbox{wrongPredictionColor}{0.97} & 0.94 & \colorbox{wrongPredictionColor}{0.95} & 1.0 \\

\midrule
$\datasetLMOrderWikipedia$ & $\modelPhiThreeFiveInstruct$ & $\estimatorNoReject$ & \colorbox{wrongPredictionColor}{0.82} & 0.69 & \colorbox{wrongPredictionColor}{0.91} & 0.27 & \colorbox{wrongPredictionColor}{0.96} & 0.59 & \colorbox{wrongPredictionColor}{0.67} & 0.37 & \colorbox{wrongPredictionColor}{0.85} & 0.96 \\
$\datasetLMOrderWikipedia$ & $\modelPhiThreeFiveInstructFinetuned$ & $\estimatorNoReject$ & \colorbox{wrongPredictionColor}{0.56} & 0.080 & \colorbox{wrongPredictionColor}{0.95} & 0.92 & \colorbox{wrongPredictionColor}{0.51} & 0.087 & \colorbox{wrongPredictionColor}{0.96} & 0.91 & \colorbox{wrongPredictionColor}{0.92} & 1.0 \\
$\datasetLMOrderWikipedia$ & $\modelPhiThreeFiveInstructFinetunedAdditionalRun$ & $\estimatorNoReject$ & \colorbox{wrongPredictionColor}{0.62} & 0.089 & \colorbox{wrongPredictionColor}{0.96} & 0.91 & \colorbox{wrongPredictionColor}{0.58} & 0.096 & \colorbox{wrongPredictionColor}{0.96} & 0.90 & \colorbox{wrongPredictionColor}{0.93} & 1.0 \\
$\datasetLMOrderWikipedia$ & $\modelPhiThreeFiveInstructFinetunedCM$ & $\estimatorNoReject$ & \colorbox{wrongPredictionColor}{0.65} & 0.066 & \colorbox{wrongPredictionColor}{0.97} & 0.93 & \colorbox{wrongPredictionColor}{0.59} & 0.073 & \colorbox{wrongPredictionColor}{0.98} & 0.93 & \colorbox{wrongPredictionColor}{0.95} & 1.0 \\
$\datasetLMOrderWikipedia$ & $\modelPhiThreeFiveInstructFinetunedCMAdditionalRun$ & $\estimatorNoReject$ & \colorbox{wrongPredictionColor}{0.61} & 0.095 & \colorbox{wrongPredictionColor}{0.97} & 0.90 & \colorbox{wrongPredictionColor}{0.71} & 0.081 & \colorbox{wrongPredictionColor}{0.96} & 0.92 & \colorbox{wrongPredictionColor}{0.94} & 1.0 \\

\midrule
$\datasetLMOrderWikipedia$ & $\modelGemmaFourEtwoBInstruct$ & $\estimatorNoReject$ & \colorbox{wrongPredictionColor}{0.98} & 1.0 & \colorbox{correctPredictionColor}{1.0} & 0.0020 & \colorbox{correctPredictionColor}{1.0} & 0.98 & \colorbox{wrongPredictionColor}{0.10} & 0.019 & \colorbox{wrongPredictionColor}{0.98} & 1.0 \\
$\datasetLMOrderWikipedia$ & $\modelGemmaFourEtwoBInstructFinetuned$ & $\estimatorNoReject$ & \colorbox{wrongPredictionColor}{0.61} & 0.096 & \colorbox{wrongPredictionColor}{0.94} & 0.90 & \colorbox{wrongPredictionColor}{0.52} & 0.11 & \colorbox{wrongPredictionColor}{0.96} & 0.89 & \colorbox{wrongPredictionColor}{0.91} & 1.0 \\
$\datasetLMOrderWikipedia$ & $\modelGemmaFourEtwoBInstructFinetunedAdditionalRun$ & $\estimatorNoReject$ & \colorbox{wrongPredictionColor}{0.46} & 0.065 & \colorbox{wrongPredictionColor}{0.94} & 0.94 & \colorbox{wrongPredictionColor}{0.34} & 0.089 & \colorbox{wrongPredictionColor}{0.96} & 0.91 & \colorbox{wrongPredictionColor}{0.91} & 1.0 \\
$\datasetLMOrderWikipedia$ & $\modelGemmaFourEtwoBInstructFinetunedCM$ & $\estimatorNoReject$ & \colorbox{wrongPredictionColor}{0.68} & 0.064 & \colorbox{wrongPredictionColor}{0.97} & 0.94 & \colorbox{wrongPredictionColor}{0.57} & 0.075 & \colorbox{wrongPredictionColor}{0.98} & 0.92 & \colorbox{wrongPredictionColor}{0.95} & 1.0 \\
$\datasetLMOrderWikipedia$ & $\modelGemmaFourEtwoBInstructFinetunedCMAdditionalRun$ & $\estimatorNoReject$ & \colorbox{wrongPredictionColor}{0.50} & 0.066 & \colorbox{wrongPredictionColor}{0.97} & 0.93 & \colorbox{wrongPredictionColor}{0.56} & 0.059 & \colorbox{wrongPredictionColor}{0.97} & 0.94 & \colorbox{wrongPredictionColor}{0.94} & 1.0 \\

\midrule
$\datasetLMOrderWikipediaCVS$ & $\modelPhiThreeFiveInstruct$ & $\estimatorNoReject$ & \colorbox{wrongPredictionColor}{0.96} & 0.81 & \colorbox{wrongPredictionColor}{0.30} & 0.15 & \colorbox{wrongPredictionColor}{0.88} & 0.88 & \colorbox{wrongPredictionColor}{0.55} & 0.078 & \colorbox{wrongPredictionColor}{0.86} & 0.96 \\
$\datasetLMOrderWikipediaCVS$ & $\modelPhiThreeFiveInstructFinetuned$ & $\estimatorNoReject$ & \colorbox{wrongPredictionColor}{0.35} & 0.12 & \colorbox{wrongPredictionColor}{0.97} & 0.89 & \colorbox{wrongPredictionColor}{0.57} & 0.070 & \colorbox{wrongPredictionColor}{0.92} & 0.93 & \colorbox{wrongPredictionColor}{0.89} & 1.0 \\
$\datasetLMOrderWikipediaCVS$ & $\modelPhiThreeFiveInstructFinetunedAdditionalRun$ & $\estimatorNoReject$ & \colorbox{wrongPredictionColor}{0.44} & 0.12 & \colorbox{wrongPredictionColor}{0.94} & 0.88 & \colorbox{wrongPredictionColor}{0.50} & 0.10 & \colorbox{wrongPredictionColor}{0.93} & 0.90 & \colorbox{wrongPredictionColor}{0.88} & 1.0 \\
$\datasetLMOrderWikipediaCVS$ & $\modelPhiThreeFiveInstructFinetunedCM$ & $\estimatorNoReject$ & \colorbox{wrongPredictionColor}{0.41} & 0.093 & \colorbox{wrongPredictionColor}{0.99} & 0.91 & \colorbox{wrongPredictionColor}{0.74} & 0.052 & \colorbox{wrongPredictionColor}{0.94} & 0.95 & \colorbox{wrongPredictionColor}{0.93} & 1.0 \\
$\datasetLMOrderWikipediaCVS$ & $\modelPhiThreeFiveInstructFinetunedCMAdditionalRun$ & $\estimatorNoReject$ & \colorbox{wrongPredictionColor}{0.46} & 0.12 & \colorbox{wrongPredictionColor}{0.98} & 0.87 & \colorbox{wrongPredictionColor}{0.75} & 0.076 & \colorbox{wrongPredictionColor}{0.93} & 0.92 & \colorbox{wrongPredictionColor}{0.91} & 1.0 \\

\midrule
$\datasetLMOrderWikipediaCVS$ & $\modelGemmaFourEtwoBInstruct$ & $\estimatorNoReject$ & \colorbox{wrongPredictionColor}{0.93} & 1.0 & \colorbox{correctPredictionColor}{1.0} & 0.0015 & \colorbox{correctPredictionColor}{1.0} & 0.93 & \colorbox{wrongPredictionColor}{0.020} & 0.074 & \colorbox{wrongPredictionColor}{0.93} & 1.0 \\
$\datasetLMOrderWikipediaCVS$ & $\modelGemmaFourEtwoBInstructFinetuned$ & $\estimatorNoReject$ & \colorbox{wrongPredictionColor}{0.31} & 0.14 & \colorbox{wrongPredictionColor}{0.95} & 0.86 & \colorbox{wrongPredictionColor}{0.48} & 0.088 & \colorbox{wrongPredictionColor}{0.90} & 0.91 & \colorbox{wrongPredictionColor}{0.86} & 1.0 \\
$\datasetLMOrderWikipediaCVS$ & $\modelGemmaFourEtwoBInstructFinetunedAdditionalRun$ & $\estimatorNoReject$ & \colorbox{wrongPredictionColor}{0.33} & 0.11 & \colorbox{wrongPredictionColor}{0.93} & 0.89 & \colorbox{wrongPredictionColor}{0.36} & 0.099 & \colorbox{wrongPredictionColor}{0.92} & 0.90 & \colorbox{wrongPredictionColor}{0.86} & 1.0 \\
$\datasetLMOrderWikipediaCVS$ & $\modelGemmaFourEtwoBInstructFinetunedCM$ & $\estimatorNoReject$ & \colorbox{wrongPredictionColor}{0.44} & 0.091 & \colorbox{wrongPredictionColor}{0.97} & 0.91 & \colorbox{wrongPredictionColor}{0.61} & 0.065 & \colorbox{wrongPredictionColor}{0.95} & 0.93 & \colorbox{wrongPredictionColor}{0.92} & 1.0 \\
$\datasetLMOrderWikipediaCVS$ & $\modelGemmaFourEtwoBInstructFinetunedCMAdditionalRun$ & $\estimatorNoReject$ & \colorbox{wrongPredictionColor}{0.28} & 0.088 & \colorbox{wrongPredictionColor}{0.98} & 0.91 & \colorbox{wrongPredictionColor}{0.64} & 0.039 & \colorbox{wrongPredictionColor}{0.93} & 0.96 & \colorbox{wrongPredictionColor}{0.92} & 1.0 \\

\midrule
$\datasetLMOrderHarvard$ & $\modelPhiThreeFiveInstruct$ & $\sdmHR$ & \colorbox{correctPredictionColor}{\allRejected} & 0.0 & \colorbox{correctPredictionColor}{\allRejected} & 0.0 & \colorbox{correctPredictionColor}{\allRejected} & 0.0 & \colorbox{correctPredictionColor}{\allRejected} & 0.0 & \colorbox{correctPredictionColor}{\allRejected} & 0.0 \\
$\datasetLMOrderHarvard$ & $\modelPhiThreeFiveInstructFinetuned$ & $\sdmHR$ & \colorbox{correctPredictionColor}{\allRejected} & 0.0 & \colorbox{correctPredictionColor}{\allRejected} & 0.0 & \colorbox{correctPredictionColor}{\allRejected} & 0.0 & \colorbox{correctPredictionColor}{\allRejected} & 0.0 & \colorbox{correctPredictionColor}{\allRejected} & 0.0 \\
$\datasetLMOrderHarvard$ & $\modelPhiThreeFiveInstructFinetunedAdditionalRun$ & $\sdmHR$ & \colorbox{correctPredictionColor}{\allRejected} & 0.0 & \colorbox{correctPredictionColor}{\allRejected} & 0.0 & \colorbox{correctPredictionColor}{\allRejected} & 0.0 & \colorbox{correctPredictionColor}{\allRejected} & 0.0 & \colorbox{correctPredictionColor}{\allRejected} & 0.0 \\
$\datasetLMOrderHarvard$ & $\modelPhiThreeFiveInstructFinetunedCM$ & $\sdmHR$ & \colorbox{wrongPredictionColor}{0.0} & 0.0025 & \colorbox{correctPredictionColor}{1.0} & 0.11 & \colorbox{correctPredictionColor}{\allRejected} & 0.0 & \colorbox{wrongPredictionColor}{0.94} & 0.11 & \colorbox{wrongPredictionColor}{0.94} & 0.11 \\
$\datasetLMOrderHarvard$ & $\modelPhiThreeFiveInstructFinetunedCMAdditionalRun$ & $\sdmHR$ & \colorbox{wrongPredictionColor}{0.0} & 0.0012 & \colorbox{correctPredictionColor}{1.0} & 0.26 & \colorbox{correctPredictionColor}{\allRejected} & 0.0 & \colorbox{correctPredictionColor}{1.0} & 0.26 & \colorbox{correctPredictionColor}{1.0} & 0.26 \\

\midrule
$\datasetLMOrderHarvard$ & $\modelGemmaFourEtwoBInstruct$ & $\sdmHR$ & \colorbox{correctPredictionColor}{1.0} & 0.63 & \colorbox{correctPredictionColor}{\allRejected} & 0.0 & \colorbox{correctPredictionColor}{1.0} & 0.63 & \colorbox{correctPredictionColor}{\allRejected} & 0.0 & \colorbox{correctPredictionColor}{1.0} & 0.63 \\
$\datasetLMOrderHarvard$ & $\modelGemmaFourEtwoBInstructFinetuned$ & $\sdmHR$ & \colorbox{correctPredictionColor}{\allRejected} & 0.0 & \colorbox{correctPredictionColor}{\allRejected} & 0.0 & \colorbox{correctPredictionColor}{\allRejected} & 0.0 & \colorbox{correctPredictionColor}{\allRejected} & 0.0 & \colorbox{correctPredictionColor}{\allRejected} & 0.0 \\
$\datasetLMOrderHarvard$ & $\modelGemmaFourEtwoBInstructFinetunedAdditionalRun$ & $\sdmHR$ & \colorbox{correctPredictionColor}{\allRejected} & 0.0 & \colorbox{correctPredictionColor}{\allRejected} & 0.0 & \colorbox{correctPredictionColor}{\allRejected} & 0.0 & \colorbox{correctPredictionColor}{\allRejected} & 0.0 & \colorbox{correctPredictionColor}{\allRejected} & 0.0 \\
$\datasetLMOrderHarvard$ & $\modelGemmaFourEtwoBInstructFinetunedCM$ & $\sdmHR$ & \colorbox{correctPredictionColor}{1.0} & 0.0014 & \colorbox{correctPredictionColor}{1.0} & 0.035 & \colorbox{correctPredictionColor}{1.0} & 0.0014 & \colorbox{correctPredictionColor}{1.0} & 0.035 & \colorbox{correctPredictionColor}{1.0} & 0.037 \\
$\datasetLMOrderHarvard$ & $\modelGemmaFourEtwoBInstructFinetunedCMAdditionalRun$ & $\sdmHR$ & \colorbox{correctPredictionColor}{1.0} & 0.00023 & \colorbox{correctPredictionColor}{1.0} & 0.10 & \colorbox{correctPredictionColor}{1.0} & 0.00023 & \colorbox{correctPredictionColor}{1.0} & 0.10 & \colorbox{correctPredictionColor}{1.0} & 0.10 \\

\midrule
$\datasetLMOrderWikipedia$ & $\modelPhiThreeFiveInstruct$ & $\sdmHR$ & \colorbox{correctPredictionColor}{\allRejected} & 0.0 & \colorbox{correctPredictionColor}{\allRejected} & 0.0 & \colorbox{correctPredictionColor}{\allRejected} & 0.0 & \colorbox{correctPredictionColor}{\allRejected} & 0.0 & \colorbox{correctPredictionColor}{\allRejected} & 0.0 \\
$\datasetLMOrderWikipedia$ & $\modelPhiThreeFiveInstructFinetuned$ & $\sdmHR$ & \colorbox{correctPredictionColor}{1.0} & 0.00017 & \colorbox{correctPredictionColor}{1.0} & 0.00025 & \colorbox{correctPredictionColor}{1.0} & 0.00017 & \colorbox{correctPredictionColor}{1.0} & 0.00025 & \colorbox{correctPredictionColor}{1.0} & 0.00042 \\
$\datasetLMOrderWikipedia$ & $\modelPhiThreeFiveInstructFinetunedAdditionalRun$ & $\sdmHR$ & \colorbox{correctPredictionColor}{\allRejected} & 0.0 & \colorbox{correctPredictionColor}{\allRejected} & 0.0 & \colorbox{correctPredictionColor}{\allRejected} & 0.0 & \colorbox{correctPredictionColor}{\allRejected} & 0.0 & \colorbox{correctPredictionColor}{\allRejected} & 0.0 \\
$\datasetLMOrderWikipedia$ & $\modelPhiThreeFiveInstructFinetunedCM$ & $\sdmHR$ & \colorbox{wrongPredictionColor}{0.72} & 0.0015 & \colorbox{correctPredictionColor}{1.0} & 0.091 & \colorbox{correctPredictionColor}{1.0} & 0.0010 & \colorbox{correctPredictionColor}{1.0} & 0.091 & \colorbox{correctPredictionColor}{1.0} & 0.092 \\
$\datasetLMOrderWikipedia$ & $\modelPhiThreeFiveInstructFinetunedCMAdditionalRun$ & $\sdmHR$ & \colorbox{wrongPredictionColor}{0.91} & 0.0026 & \colorbox{correctPredictionColor}{1.0} & 0.079 & \colorbox{correctPredictionColor}{1.0} & 0.0023 & \colorbox{correctPredictionColor}{1.0} & 0.079 & \colorbox{correctPredictionColor}{1.0} & 0.081 \\

\midrule
$\datasetLMOrderWikipedia$ & $\modelGemmaFourEtwoBInstruct$ & $\sdmHR$ & \colorbox{correctPredictionColor}{1.0} & 0.63 & \colorbox{correctPredictionColor}{1.0} & 0.0010 & \colorbox{correctPredictionColor}{1.0} & 0.63 & \colorbox{wrongPredictionColor}{0.61} & 0.0017 & \colorbox{correctPredictionColor}{1.0} & 0.63 \\
$\datasetLMOrderWikipedia$ & $\modelGemmaFourEtwoBInstructFinetuned$ & $\sdmHR$ & \colorbox{correctPredictionColor}{\allRejected} & 0.0 & \colorbox{correctPredictionColor}{\allRejected} & 0.0 & \colorbox{correctPredictionColor}{\allRejected} & 0.0 & \colorbox{correctPredictionColor}{\allRejected} & 0.0 & \colorbox{correctPredictionColor}{\allRejected} & 0.0 \\
$\datasetLMOrderWikipedia$ & $\modelGemmaFourEtwoBInstructFinetunedAdditionalRun$ & $\sdmHR$ & \colorbox{correctPredictionColor}{\allRejected} & 0.0 & \colorbox{correctPredictionColor}{\allRejected} & 0.0 & \colorbox{correctPredictionColor}{\allRejected} & 0.0 & \colorbox{correctPredictionColor}{\allRejected} & 0.0 & \colorbox{correctPredictionColor}{\allRejected} & 0.0 \\
$\datasetLMOrderWikipedia$ & $\modelGemmaFourEtwoBInstructFinetunedCM$ & $\sdmHR$ & \colorbox{wrongPredictionColor}{0.92} & 0.00058 & \colorbox{correctPredictionColor}{0.99} & 0.049 & \colorbox{wrongPredictionColor}{0.67} & 0.00083 & \colorbox{correctPredictionColor}{1.0} & 0.049 & \colorbox{correctPredictionColor}{0.99} & 0.049 \\
$\datasetLMOrderWikipedia$ & $\modelGemmaFourEtwoBInstructFinetunedCMAdditionalRun$ & $\sdmHR$ & \colorbox{correctPredictionColor}{1.0} & 0.00092 & \colorbox{correctPredictionColor}{1.0} & 0.035 & \colorbox{correctPredictionColor}{1.0} & 0.00092 & \colorbox{correctPredictionColor}{1.0} & 0.035 & \colorbox{correctPredictionColor}{1.0} & 0.036 \\

\midrule
$\datasetLMOrderWikipediaCVS$ & $\modelPhiThreeFiveInstruct$ & $\sdmHR$ & \colorbox{correctPredictionColor}{\allRejected} & 0.0 & \colorbox{correctPredictionColor}{\allRejected} & 0.0 & \colorbox{correctPredictionColor}{\allRejected} & 0.0 & \colorbox{correctPredictionColor}{\allRejected} & 0.0 & \colorbox{correctPredictionColor}{\allRejected} & 0.0 \\
$\datasetLMOrderWikipediaCVS$ & $\modelPhiThreeFiveInstructFinetuned$ & $\sdmHR$ & \colorbox{correctPredictionColor}{\allRejected} & 0.0 & \colorbox{correctPredictionColor}{\allRejected} & 0.0 & \colorbox{correctPredictionColor}{\allRejected} & 0.0 & \colorbox{correctPredictionColor}{\allRejected} & 0.0 & \colorbox{correctPredictionColor}{\allRejected} & 0.0 \\
$\datasetLMOrderWikipediaCVS$ & $\modelPhiThreeFiveInstructFinetunedAdditionalRun$ & $\sdmHR$ & \colorbox{correctPredictionColor}{\allRejected} & 0.0 & \colorbox{correctPredictionColor}{\allRejected} & 0.0 & \colorbox{correctPredictionColor}{\allRejected} & 0.0 & \colorbox{correctPredictionColor}{\allRejected} & 0.0 & \colorbox{correctPredictionColor}{\allRejected} & 0.0 \\
$\datasetLMOrderWikipediaCVS$ & $\modelPhiThreeFiveInstructFinetunedCM$ & $\sdmHR$ & \colorbox{correctPredictionColor}{\allRejected} & 0.0 & \colorbox{correctPredictionColor}{\allRejected} & 0.0 & \colorbox{correctPredictionColor}{\allRejected} & 0.0 & \colorbox{correctPredictionColor}{\allRejected} & 0.0 & \colorbox{correctPredictionColor}{\allRejected} & 0.0 \\
$\datasetLMOrderWikipediaCVS$ & $\modelPhiThreeFiveInstructFinetunedCMAdditionalRun$ & $\sdmHR$ & \colorbox{correctPredictionColor}{1.0} & 0.00050 & \colorbox{correctPredictionColor}{\allRejected} & 0.0 & \colorbox{correctPredictionColor}{1.0} & 0.00050 & \colorbox{correctPredictionColor}{\allRejected} & 0.0 & \colorbox{correctPredictionColor}{1.0} & 0.00050 \\

\midrule
$\datasetLMOrderWikipediaCVS$ & $\modelGemmaFourEtwoBInstruct$ & $\sdmHR$ & \colorbox{correctPredictionColor}{1.0} & 0.20 & \colorbox{correctPredictionColor}{\allRejected} & 0.0 & \colorbox{correctPredictionColor}{1.0} & 0.20 & \colorbox{wrongPredictionColor}{0.0} & 0.00042 & \colorbox{correctPredictionColor}{1.0} & 0.20 \\
$\datasetLMOrderWikipediaCVS$ & $\modelGemmaFourEtwoBInstructFinetuned$ & $\sdmHR$ & \colorbox{correctPredictionColor}{\allRejected} & 0.0 & \colorbox{correctPredictionColor}{\allRejected} & 0.0 & \colorbox{correctPredictionColor}{\allRejected} & 0.0 & \colorbox{correctPredictionColor}{\allRejected} & 0.0 & \colorbox{correctPredictionColor}{\allRejected} & 0.0 \\
$\datasetLMOrderWikipediaCVS$ & $\modelGemmaFourEtwoBInstructFinetunedAdditionalRun$ & $\sdmHR$ & \colorbox{correctPredictionColor}{\allRejected} & 0.0 & \colorbox{correctPredictionColor}{\allRejected} & 0.0 & \colorbox{correctPredictionColor}{\allRejected} & 0.0 & \colorbox{correctPredictionColor}{\allRejected} & 0.0 & \colorbox{correctPredictionColor}{\allRejected} & 0.0 \\
$\datasetLMOrderWikipediaCVS$ & $\modelGemmaFourEtwoBInstructFinetunedCM$ & $\sdmHR$ & \colorbox{correctPredictionColor}{\allRejected} & 0.0 & \colorbox{correctPredictionColor}{1.0} & 0.0024 & \colorbox{correctPredictionColor}{\allRejected} & 0.0 & \colorbox{correctPredictionColor}{1.0} & 0.0024 & \colorbox{correctPredictionColor}{1.0} & 0.0024 \\
$\datasetLMOrderWikipediaCVS$ & $\modelGemmaFourEtwoBInstructFinetunedCMAdditionalRun$ & $\sdmHR$ & \colorbox{correctPredictionColor}{\allRejected} & 0.0 & \colorbox{correctPredictionColor}{1.0} & 0.00017 & \colorbox{correctPredictionColor}{\allRejected} & 0.0 & \colorbox{correctPredictionColor}{1.0} & 0.00017 & \colorbox{correctPredictionColor}{1.0} & 0.00017 \\

    \bottomrule
  \end{tabular}
  }  
    \caption{Comparison of estimators for the word ordering task, with \colorbox{correctPredictionColor}{$\alpha$}$=0.99$, averaged over 6 independently calibrated $\sdm$ estimators. \colorbox{correctPredictionColor}{\allRejected} indicates all predictions were rejected across all estimators, which is preferred over falling \colorbox{wrongPredictionColor}{under} the expected accuracy. (Highlighting is applied prior to rounding.)
  $n=|\text{Admitted}|$, the count of non-rejected documents. Here, $\hat{y}=\argmax \vz'$, the document-level prediction from the $\sdm$ activation layer, with $y=r(\vs^+, \hat{\vs})$. For reference, we include the output from the $\sdm$ activation layer without any selective filtering as the $\estimatorNoReject$ estimator. Unlike the results on static datasets in Table~\ref{tab:experiments-verification-results-static}, in this case the true class proportions are model dependent since they are derived from test-time generations via $y=r(\vs^+, \hat{\vs})$. The base models are notably skewed to negative generations. For reference, the $\modelPhiThreeFiveInstructFinetunedCM$ row for the $\datasetLMOrderHarvard$ dataset had the following raw counts across the six $\sdmHR$ estimators for the marginal accuracy: $\frac{77}{79}, \frac{142}{144}, \frac{122}{124}, \frac{55}{57}, \frac{3}{4}, \frac{61}{63}$. These errors come from only two distinct responses. The LM generation ``Soap can wash away most dirt.'' is a grammatical alternative to the reference ``Soap can wash most dirt away.'', although it relocates the period. The other LM response deviates from the reference ``The term ended in late june that year.'' only by capitalizing ``June'' (which is incorrectly cased in the reference). Both fail the strict exact reference matching, but are arguably within the aleatoric (irreducible) error of the task.} %
  \label{tab:experiments-verification-results} 
\end{table*}

\begin{table*}
  \centering
  \setlength{\fboxsep}{1pt}
  \resizebox{0.82\textwidth}{!}{%
  \begin{tabular}{l l  l c c c c c c c c c l }
    \toprule

    & & &  \multicolumn{4}{c}{Class-conditional} & \multicolumn{4}{c}{Prediction-conditional}  & \multicolumn{2}{c}{Marginal} \\
    & & &  \multicolumn{2}{c}{$y=0$} & \multicolumn{2}{c}{$y=1$} & \multicolumn{2}{c}{$\hat{y}=0$} & \multicolumn{2}{c}{$\hat{y}=1$} & \multicolumn{2}{c}{$y\in \{0,1\}$} \\
    \cmidrule(r){4-5} \cmidrule(r){6-7} \cmidrule(r){8-9} \cmidrule(r){10-11} \cmidrule(r){12-13}
    Dataset   & Model & Estimator & \textsc{Acc.}& $\frac{n}{|\testSplit|}$ & \textsc{Acc.} & $\frac{n}{|\testSplit|}$ & \textsc{Acc.} & $\frac{n}{|\testSplit|}$ & \textsc{Acc.} & $\frac{n}{|\testSplit|}$ & \textsc{Acc.} & $\frac{n}{|\testSplit|}$\\
   
   \midrule
  
$\datasetLMOrderStaticHarvard$ & $\modelPhiThreeFiveInstruct$ & $\estimatorNoReject$ & \colorbox{wrongPredictionColor}{0.98} & 0.50 & \colorbox{wrongPredictionColor}{0.98} & 0.50 & \colorbox{wrongPredictionColor}{0.98} & 0.50 & \colorbox{wrongPredictionColor}{0.98} & 0.50 & \colorbox{wrongPredictionColor}{0.98} & 1.0 \\
$\datasetLMOrderStaticHarvard$ & $\modelPhiThreeFiveInstructFinetuned$ & $\estimatorNoReject$ & \colorbox{wrongPredictionColor}{0.99} & 0.50 & \colorbox{wrongPredictionColor}{0.98} & 0.50 & \colorbox{wrongPredictionColor}{0.98} & 0.50 & \colorbox{wrongPredictionColor}{0.99} & 0.50 & \colorbox{wrongPredictionColor}{0.99} & 1.0 \\
$\datasetLMOrderStaticHarvard$ & $\modelPhiThreeFiveInstructFinetunedAdditionalRun$ & $\estimatorNoReject$ & \colorbox{wrongPredictionColor}{0.99} & 0.50 & \colorbox{wrongPredictionColor}{0.98} & 0.50 & \colorbox{wrongPredictionColor}{0.98} & 0.50 & \colorbox{wrongPredictionColor}{0.99} & 0.50 & \colorbox{wrongPredictionColor}{0.98} & 1.0 \\
$\datasetLMOrderStaticHarvard$ & $\modelPhiThreeFiveInstructFinetunedCM$ & $\estimatorNoReject$ & \colorbox{wrongPredictionColor}{0.99} & 0.50 & \colorbox{wrongPredictionColor}{0.97} & 0.50 & \colorbox{wrongPredictionColor}{0.97} & 0.51 & \colorbox{wrongPredictionColor}{0.99} & 0.49 & \colorbox{wrongPredictionColor}{0.98} & 1.0 \\
$\datasetLMOrderStaticHarvard$ & $\modelPhiThreeFiveInstructFinetunedCMAdditionalRun$ & $\estimatorNoReject$ & \colorbox{wrongPredictionColor}{0.99} & 0.50 & \colorbox{wrongPredictionColor}{0.98} & 0.50 & \colorbox{wrongPredictionColor}{0.98} & 0.50 & \colorbox{wrongPredictionColor}{0.99} & 0.50 & \colorbox{wrongPredictionColor}{0.98} & 1.0 \\

\midrule
$\datasetLMOrderStaticHarvard$ & $\modelGemmaFourEtwoBInstruct$ & $\estimatorNoReject$ & \colorbox{wrongPredictionColor}{0.97} & 0.50 & \colorbox{wrongPredictionColor}{0.97} & 0.50 & \colorbox{wrongPredictionColor}{0.97} & 0.50 & \colorbox{wrongPredictionColor}{0.97} & 0.50 & \colorbox{wrongPredictionColor}{0.97} & 1.0 \\
$\datasetLMOrderStaticHarvard$ & $\modelGemmaFourEtwoBInstructFinetuned$ & $\estimatorNoReject$ & \colorbox{wrongPredictionColor}{0.97} & 0.50 & \colorbox{wrongPredictionColor}{0.98} & 0.50 & \colorbox{wrongPredictionColor}{0.98} & 0.50 & \colorbox{wrongPredictionColor}{0.97} & 0.50 & \colorbox{wrongPredictionColor}{0.98} & 1.0 \\
$\datasetLMOrderStaticHarvard$ & $\modelGemmaFourEtwoBInstructFinetunedAdditionalRun$ & $\estimatorNoReject$ & \colorbox{wrongPredictionColor}{0.97} & 0.50 & \colorbox{wrongPredictionColor}{0.93} & 0.50 & \colorbox{wrongPredictionColor}{0.94} & 0.52 & \colorbox{wrongPredictionColor}{0.97} & 0.48 & \colorbox{wrongPredictionColor}{0.95} & 1.0 \\
$\datasetLMOrderStaticHarvard$ & $\modelGemmaFourEtwoBInstructFinetunedCM$ & $\estimatorNoReject$ & \colorbox{wrongPredictionColor}{0.99} & 0.50 & \colorbox{wrongPredictionColor}{0.94} & 0.50 & \colorbox{wrongPredictionColor}{0.94} & 0.53 & \colorbox{wrongPredictionColor}{0.99} & 0.47 & \colorbox{wrongPredictionColor}{0.96} & 1.0 \\
$\datasetLMOrderStaticHarvard$ & $\modelGemmaFourEtwoBInstructFinetunedCMAdditionalRun$ & $\estimatorNoReject$ & \colorbox{correctPredictionColor}{0.99} & 0.50 & \colorbox{wrongPredictionColor}{0.94} & 0.50 & \colorbox{wrongPredictionColor}{0.95} & 0.52 & \colorbox{correctPredictionColor}{0.99} & 0.48 & \colorbox{wrongPredictionColor}{0.97} & 1.0 \\

\midrule
$\datasetLMOrderStaticWikipedia$ & $\modelPhiThreeFiveInstruct$ & $\estimatorNoReject$ & \colorbox{wrongPredictionColor}{0.97} & 0.50 & \colorbox{wrongPredictionColor}{0.97} & 0.50 & \colorbox{wrongPredictionColor}{0.97} & 0.50 & \colorbox{wrongPredictionColor}{0.97} & 0.50 & \colorbox{wrongPredictionColor}{0.97} & 1.0 \\
$\datasetLMOrderStaticWikipedia$ & $\modelPhiThreeFiveInstructFinetuned$ & $\estimatorNoReject$ & \colorbox{wrongPredictionColor}{0.98} & 0.50 & \colorbox{wrongPredictionColor}{0.96} & 0.50 & \colorbox{wrongPredictionColor}{0.96} & 0.51 & \colorbox{wrongPredictionColor}{0.98} & 0.49 & \colorbox{wrongPredictionColor}{0.97} & 1.0 \\
$\datasetLMOrderStaticWikipedia$ & $\modelPhiThreeFiveInstructFinetunedAdditionalRun$ & $\estimatorNoReject$ & \colorbox{wrongPredictionColor}{0.98} & 0.50 & \colorbox{wrongPredictionColor}{0.97} & 0.50 & \colorbox{wrongPredictionColor}{0.97} & 0.50 & \colorbox{wrongPredictionColor}{0.98} & 0.50 & \colorbox{wrongPredictionColor}{0.97} & 1.0 \\
$\datasetLMOrderStaticWikipedia$ & $\modelPhiThreeFiveInstructFinetunedCM$ & $\estimatorNoReject$ & \colorbox{wrongPredictionColor}{0.99} & 0.50 & \colorbox{wrongPredictionColor}{0.98} & 0.50 & \colorbox{wrongPredictionColor}{0.98} & 0.51 & \colorbox{wrongPredictionColor}{0.99} & 0.49 & \colorbox{wrongPredictionColor}{0.98} & 1.0 \\
$\datasetLMOrderStaticWikipedia$ & $\modelPhiThreeFiveInstructFinetunedCMAdditionalRun$ & $\estimatorNoReject$ & \colorbox{wrongPredictionColor}{0.98} & 0.50 & \colorbox{wrongPredictionColor}{0.99} & 0.50 & \colorbox{wrongPredictionColor}{0.99} & 0.50 & \colorbox{wrongPredictionColor}{0.98} & 0.50 & \colorbox{wrongPredictionColor}{0.98} & 1.0 \\

\midrule
$\datasetLMOrderStaticWikipedia$ & $\modelGemmaFourEtwoBInstruct$ & $\estimatorNoReject$ & \colorbox{wrongPredictionColor}{0.94} & 0.50 & \colorbox{wrongPredictionColor}{0.94} & 0.50 & \colorbox{wrongPredictionColor}{0.94} & 0.50 & \colorbox{wrongPredictionColor}{0.94} & 0.50 & \colorbox{wrongPredictionColor}{0.94} & 1.0 \\
$\datasetLMOrderStaticWikipedia$ & $\modelGemmaFourEtwoBInstructFinetuned$ & $\estimatorNoReject$ & \colorbox{wrongPredictionColor}{0.95} & 0.50 & \colorbox{wrongPredictionColor}{0.95} & 0.50 & \colorbox{wrongPredictionColor}{0.95} & 0.50 & \colorbox{wrongPredictionColor}{0.95} & 0.50 & \colorbox{wrongPredictionColor}{0.95} & 1.0 \\
$\datasetLMOrderStaticWikipedia$ & $\modelGemmaFourEtwoBInstructFinetunedAdditionalRun$ & $\estimatorNoReject$ & \colorbox{wrongPredictionColor}{0.93} & 0.50 & \colorbox{wrongPredictionColor}{0.95} & 0.50 & \colorbox{wrongPredictionColor}{0.95} & 0.49 & \colorbox{wrongPredictionColor}{0.93} & 0.51 & \colorbox{wrongPredictionColor}{0.94} & 1.0 \\
$\datasetLMOrderStaticWikipedia$ & $\modelGemmaFourEtwoBInstructFinetunedCM$ & $\estimatorNoReject$ & \colorbox{wrongPredictionColor}{0.98} & 0.50 & \colorbox{wrongPredictionColor}{0.98} & 0.50 & \colorbox{wrongPredictionColor}{0.98} & 0.50 & \colorbox{wrongPredictionColor}{0.98} & 0.50 & \colorbox{wrongPredictionColor}{0.98} & 1.0 \\
$\datasetLMOrderStaticWikipedia$ & $\modelGemmaFourEtwoBInstructFinetunedCMAdditionalRun$ & $\estimatorNoReject$ & \colorbox{wrongPredictionColor}{0.98} & 0.50 & \colorbox{wrongPredictionColor}{0.98} & 0.50 & \colorbox{wrongPredictionColor}{0.98} & 0.50 & \colorbox{wrongPredictionColor}{0.98} & 0.50 & \colorbox{wrongPredictionColor}{0.98} & 1.0 \\

\midrule
$\datasetLMOrderStaticWikipediaCVS$ & $\modelPhiThreeFiveInstruct$ & $\estimatorNoReject$ & \colorbox{wrongPredictionColor}{0.95} & 0.50 & \colorbox{wrongPredictionColor}{0.96} & 0.50 & \colorbox{wrongPredictionColor}{0.96} & 0.50 & \colorbox{wrongPredictionColor}{0.95} & 0.50 & \colorbox{wrongPredictionColor}{0.96} & 1.0 \\
$\datasetLMOrderStaticWikipediaCVS$ & $\modelPhiThreeFiveInstructFinetuned$ & $\estimatorNoReject$ & \colorbox{wrongPredictionColor}{0.94} & 0.50 & \colorbox{wrongPredictionColor}{0.98} & 0.50 & \colorbox{wrongPredictionColor}{0.98} & 0.48 & \colorbox{wrongPredictionColor}{0.94} & 0.52 & \colorbox{wrongPredictionColor}{0.96} & 1.0 \\
$\datasetLMOrderStaticWikipediaCVS$ & $\modelPhiThreeFiveInstructFinetunedAdditionalRun$ & $\estimatorNoReject$ & \colorbox{wrongPredictionColor}{0.97} & 0.50 & \colorbox{wrongPredictionColor}{0.96} & 0.50 & \colorbox{wrongPredictionColor}{0.96} & 0.50 & \colorbox{wrongPredictionColor}{0.97} & 0.50 & \colorbox{wrongPredictionColor}{0.97} & 1.0 \\
$\datasetLMOrderStaticWikipediaCVS$ & $\modelPhiThreeFiveInstructFinetunedCM$ & $\estimatorNoReject$ & \colorbox{wrongPredictionColor}{0.98} & 0.50 & \colorbox{wrongPredictionColor}{0.98} & 0.50 & \colorbox{wrongPredictionColor}{0.98} & 0.50 & \colorbox{wrongPredictionColor}{0.98} & 0.50 & \colorbox{wrongPredictionColor}{0.98} & 1.0 \\
$\datasetLMOrderStaticWikipediaCVS$ & $\modelPhiThreeFiveInstructFinetunedCMAdditionalRun$ & $\estimatorNoReject$ & \colorbox{wrongPredictionColor}{0.98} & 0.50 & \colorbox{wrongPredictionColor}{0.98} & 0.50 & \colorbox{wrongPredictionColor}{0.98} & 0.50 & \colorbox{wrongPredictionColor}{0.98} & 0.50 & \colorbox{wrongPredictionColor}{0.98} & 1.0 \\

\midrule
$\datasetLMOrderStaticWikipediaCVS$ & $\modelGemmaFourEtwoBInstruct$ & $\estimatorNoReject$ & \colorbox{wrongPredictionColor}{0.88} & 0.50 & \colorbox{wrongPredictionColor}{0.91} & 0.50 & \colorbox{wrongPredictionColor}{0.91} & 0.48 & \colorbox{wrongPredictionColor}{0.88} & 0.52 & \colorbox{wrongPredictionColor}{0.89} & 1.0 \\
$\datasetLMOrderStaticWikipediaCVS$ & $\modelGemmaFourEtwoBInstructFinetuned$ & $\estimatorNoReject$ & \colorbox{wrongPredictionColor}{0.89} & 0.50 & \colorbox{wrongPredictionColor}{0.98} & 0.50 & \colorbox{wrongPredictionColor}{0.98} & 0.46 & \colorbox{wrongPredictionColor}{0.90} & 0.54 & \colorbox{wrongPredictionColor}{0.94} & 1.0 \\
$\datasetLMOrderStaticWikipediaCVS$ & $\modelGemmaFourEtwoBInstructFinetunedAdditionalRun$ & $\estimatorNoReject$ & \colorbox{wrongPredictionColor}{0.88} & 0.50 & \colorbox{wrongPredictionColor}{0.94} & 0.50 & \colorbox{wrongPredictionColor}{0.93} & 0.47 & \colorbox{wrongPredictionColor}{0.88} & 0.53 & \colorbox{wrongPredictionColor}{0.91} & 1.0 \\
$\datasetLMOrderStaticWikipediaCVS$ & $\modelGemmaFourEtwoBInstructFinetunedCM$ & $\estimatorNoReject$ & \colorbox{wrongPredictionColor}{0.97} & 0.50 & \colorbox{wrongPredictionColor}{0.98} & 0.50 & \colorbox{wrongPredictionColor}{0.98} & 0.49 & \colorbox{wrongPredictionColor}{0.97} & 0.51 & \colorbox{wrongPredictionColor}{0.98} & 1.0 \\
$\datasetLMOrderStaticWikipediaCVS$ & $\modelGemmaFourEtwoBInstructFinetunedCMAdditionalRun$ & $\estimatorNoReject$ & \colorbox{wrongPredictionColor}{0.98} & 0.50 & \colorbox{wrongPredictionColor}{0.99} & 0.50 & \colorbox{wrongPredictionColor}{0.99} & 0.49 & \colorbox{wrongPredictionColor}{0.98} & 0.51 & \colorbox{wrongPredictionColor}{0.98} & 1.0 \\

\midrule
$\datasetLMOrderStaticHarvard$ & $\modelPhiThreeFiveInstruct$ & $\sdmHR$ & \colorbox{correctPredictionColor}{1.0} & 0.31 & \colorbox{correctPredictionColor}{1.0} & 0.35 & \colorbox{correctPredictionColor}{1.0} & 0.31 & \colorbox{correctPredictionColor}{1.0} & 0.35 & \colorbox{correctPredictionColor}{1.0} & 0.66 \\
$\datasetLMOrderStaticHarvard$ & $\modelPhiThreeFiveInstructFinetuned$ & $\sdmHR$ & \colorbox{correctPredictionColor}{1.0} & 0.42 & \colorbox{correctPredictionColor}{1.0} & 0.37 & \colorbox{correctPredictionColor}{1.0} & 0.42 & \colorbox{correctPredictionColor}{1.0} & 0.37 & \colorbox{correctPredictionColor}{1.0} & 0.79 \\
$\datasetLMOrderStaticHarvard$ & $\modelPhiThreeFiveInstructFinetunedAdditionalRun$ & $\sdmHR$ & \colorbox{correctPredictionColor}{1.0} & 0.39 & \colorbox{correctPredictionColor}{1.0} & 0.34 & \colorbox{correctPredictionColor}{1.0} & 0.39 & \colorbox{correctPredictionColor}{1.0} & 0.34 & \colorbox{correctPredictionColor}{1.0} & 0.73 \\
$\datasetLMOrderStaticHarvard$ & $\modelPhiThreeFiveInstructFinetunedCM$ & $\sdmHR$ & \colorbox{correctPredictionColor}{1.0} & 0.42 & \colorbox{correctPredictionColor}{1.0} & 0.34 & \colorbox{correctPredictionColor}{1.0} & 0.42 & \colorbox{correctPredictionColor}{1.0} & 0.34 & \colorbox{correctPredictionColor}{1.0} & 0.76 \\
$\datasetLMOrderStaticHarvard$ & $\modelPhiThreeFiveInstructFinetunedCMAdditionalRun$ & $\sdmHR$ & \colorbox{correctPredictionColor}{1.0} & 0.40 & \colorbox{correctPredictionColor}{1.0} & 0.39 & \colorbox{correctPredictionColor}{1.0} & 0.40 & \colorbox{correctPredictionColor}{1.0} & 0.39 & \colorbox{correctPredictionColor}{1.0} & 0.79 \\

\midrule
$\datasetLMOrderStaticHarvard$ & $\modelGemmaFourEtwoBInstruct$ & $\sdmHR$ & \colorbox{correctPredictionColor}{1.0} & 0.0022 & \colorbox{correctPredictionColor}{\allRejected} & 0.0 & \colorbox{correctPredictionColor}{1.0} & 0.0022 & \colorbox{correctPredictionColor}{\allRejected} & 0.0 & \colorbox{correctPredictionColor}{1.0} & 0.0022 \\
$\datasetLMOrderStaticHarvard$ & $\modelGemmaFourEtwoBInstructFinetuned$ & $\sdmHR$ & \colorbox{correctPredictionColor}{1.0} & 0.38 & \colorbox{correctPredictionColor}{1.0} & 0.023 & \colorbox{correctPredictionColor}{1.0} & 0.38 & \colorbox{correctPredictionColor}{1.0} & 0.023 & \colorbox{correctPredictionColor}{1.0} & 0.40 \\
$\datasetLMOrderStaticHarvard$ & $\modelGemmaFourEtwoBInstructFinetunedAdditionalRun$ & $\sdmHR$ & \colorbox{correctPredictionColor}{1.0} & 0.36 & \colorbox{correctPredictionColor}{\allRejected} & 0.0 & \colorbox{correctPredictionColor}{1.0} & 0.36 & \colorbox{correctPredictionColor}{\allRejected} & 0.0 & \colorbox{correctPredictionColor}{1.0} & 0.36 \\
$\datasetLMOrderStaticHarvard$ & $\modelGemmaFourEtwoBInstructFinetunedCM$ & $\sdmHR$ & \colorbox{correctPredictionColor}{1.0} & 0.41 & \colorbox{correctPredictionColor}{1.0} & 0.33 & \colorbox{correctPredictionColor}{1.0} & 0.41 & \colorbox{correctPredictionColor}{1.0} & 0.33 & \colorbox{correctPredictionColor}{1.0} & 0.74 \\
$\datasetLMOrderStaticHarvard$ & $\modelGemmaFourEtwoBInstructFinetunedCMAdditionalRun$ & $\sdmHR$ & \colorbox{correctPredictionColor}{1.0} & 0.45 & \colorbox{correctPredictionColor}{1.0} & 0.41 & \colorbox{correctPredictionColor}{1.0} & 0.45 & \colorbox{correctPredictionColor}{1.0} & 0.41 & \colorbox{correctPredictionColor}{1.0} & 0.86 \\

\midrule
$\datasetLMOrderStaticWikipedia$ & $\modelPhiThreeFiveInstruct$ & $\sdmHR$ & \colorbox{correctPredictionColor}{1.0} & 0.34 & \colorbox{correctPredictionColor}{1.0} & 0.17 & \colorbox{correctPredictionColor}{1.0} & 0.34 & \colorbox{correctPredictionColor}{1.0} & 0.17 & \colorbox{correctPredictionColor}{1.0} & 0.52 \\
$\datasetLMOrderStaticWikipedia$ & $\modelPhiThreeFiveInstructFinetuned$ & $\sdmHR$ & \colorbox{correctPredictionColor}{1.0} & 0.38 & \colorbox{correctPredictionColor}{1.0} & 0.25 & \colorbox{correctPredictionColor}{1.0} & 0.38 & \colorbox{correctPredictionColor}{1.0} & 0.25 & \colorbox{correctPredictionColor}{1.0} & 0.63 \\
$\datasetLMOrderStaticWikipedia$ & $\modelPhiThreeFiveInstructFinetunedAdditionalRun$ & $\sdmHR$ & \colorbox{correctPredictionColor}{1.0} & 0.37 & \colorbox{correctPredictionColor}{1.0} & 0.27 & \colorbox{correctPredictionColor}{1.0} & 0.37 & \colorbox{correctPredictionColor}{1.0} & 0.27 & \colorbox{correctPredictionColor}{1.0} & 0.64 \\
$\datasetLMOrderStaticWikipedia$ & $\modelPhiThreeFiveInstructFinetunedCM$ & $\sdmHR$ & \colorbox{correctPredictionColor}{1.0} & 0.41 & \colorbox{correctPredictionColor}{1.0} & 0.31 & \colorbox{correctPredictionColor}{1.0} & 0.41 & \colorbox{correctPredictionColor}{1.0} & 0.31 & \colorbox{correctPredictionColor}{1.0} & 0.72 \\
$\datasetLMOrderStaticWikipedia$ & $\modelPhiThreeFiveInstructFinetunedCMAdditionalRun$ & $\sdmHR$ & \colorbox{correctPredictionColor}{1.0} & 0.41 & \colorbox{correctPredictionColor}{1.0} & 0.37 & \colorbox{correctPredictionColor}{1.0} & 0.41 & \colorbox{correctPredictionColor}{1.0} & 0.37 & \colorbox{correctPredictionColor}{1.0} & 0.78 \\

\midrule
$\datasetLMOrderStaticWikipedia$ & $\modelGemmaFourEtwoBInstruct$ & $\sdmHR$ & \colorbox{correctPredictionColor}{1.0} & 0.0080 & \colorbox{correctPredictionColor}{1.0} & 0.00017 & \colorbox{correctPredictionColor}{1.0} & 0.0080 & \colorbox{correctPredictionColor}{1.0} & 0.00017 & \colorbox{correctPredictionColor}{1.0} & 0.0082 \\
$\datasetLMOrderStaticWikipedia$ & $\modelGemmaFourEtwoBInstructFinetuned$ & $\sdmHR$ & \colorbox{correctPredictionColor}{1.0} & 0.38 & \colorbox{correctPredictionColor}{1.0} & 0.032 & \colorbox{correctPredictionColor}{1.0} & 0.38 & \colorbox{correctPredictionColor}{1.0} & 0.032 & \colorbox{correctPredictionColor}{1.0} & 0.41 \\
$\datasetLMOrderStaticWikipedia$ & $\modelGemmaFourEtwoBInstructFinetunedAdditionalRun$ & $\sdmHR$ & \colorbox{correctPredictionColor}{1.0} & 0.33 & \colorbox{correctPredictionColor}{1.0} & 0.0032 & \colorbox{correctPredictionColor}{1.0} & 0.33 & \colorbox{correctPredictionColor}{1.0} & 0.0032 & \colorbox{correctPredictionColor}{1.0} & 0.33 \\
$\datasetLMOrderStaticWikipedia$ & $\modelGemmaFourEtwoBInstructFinetunedCM$ & $\sdmHR$ & \colorbox{correctPredictionColor}{1.0} & 0.40 & \colorbox{correctPredictionColor}{1.0} & 0.36 & \colorbox{correctPredictionColor}{1.0} & 0.40 & \colorbox{correctPredictionColor}{1.0} & 0.36 & \colorbox{correctPredictionColor}{1.0} & 0.76 \\
$\datasetLMOrderStaticWikipedia$ & $\modelGemmaFourEtwoBInstructFinetunedCMAdditionalRun$ & $\sdmHR$ & \colorbox{correctPredictionColor}{1.0} & 0.44 & \colorbox{correctPredictionColor}{1.0} & 0.39 & \colorbox{correctPredictionColor}{1.0} & 0.44 & \colorbox{correctPredictionColor}{1.0} & 0.39 & \colorbox{correctPredictionColor}{1.0} & 0.82 \\

\midrule
$\datasetLMOrderStaticWikipediaCVS$ & $\modelPhiThreeFiveInstruct$ & $\sdmHR$ & \colorbox{correctPredictionColor}{1.0} & 0.098 & \colorbox{correctPredictionColor}{\allRejected} & 0.0 & \colorbox{correctPredictionColor}{1.0} & 0.098 & \colorbox{correctPredictionColor}{\allRejected} & 0.0 & \colorbox{correctPredictionColor}{1.0} & 0.098 \\
$\datasetLMOrderStaticWikipediaCVS$ & $\modelPhiThreeFiveInstructFinetuned$ & $\sdmHR$ & \colorbox{correctPredictionColor}{1.0} & 0.060 & \colorbox{correctPredictionColor}{1.0} & 0.020 & \colorbox{correctPredictionColor}{1.0} & 0.060 & \colorbox{correctPredictionColor}{1.0} & 0.020 & \colorbox{correctPredictionColor}{1.0} & 0.080 \\
$\datasetLMOrderStaticWikipediaCVS$ & $\modelPhiThreeFiveInstructFinetunedAdditionalRun$ & $\sdmHR$ & \colorbox{correctPredictionColor}{1.0} & 0.13 & \colorbox{correctPredictionColor}{1.0} & 0.039 & \colorbox{correctPredictionColor}{1.0} & 0.13 & \colorbox{correctPredictionColor}{1.0} & 0.039 & \colorbox{correctPredictionColor}{1.0} & 0.17 \\
$\datasetLMOrderStaticWikipediaCVS$ & $\modelPhiThreeFiveInstructFinetunedCM$ & $\sdmHR$ & \colorbox{correctPredictionColor}{1.0} & 0.13 & \colorbox{correctPredictionColor}{1.0} & 0.052 & \colorbox{correctPredictionColor}{1.0} & 0.13 & \colorbox{correctPredictionColor}{1.0} & 0.052 & \colorbox{correctPredictionColor}{1.0} & 0.18 \\
$\datasetLMOrderStaticWikipediaCVS$ & $\modelPhiThreeFiveInstructFinetunedCMAdditionalRun$ & $\sdmHR$ & \colorbox{correctPredictionColor}{1.0} & 0.24 & \colorbox{correctPredictionColor}{1.0} & 0.12 & \colorbox{correctPredictionColor}{1.0} & 0.24 & \colorbox{correctPredictionColor}{1.0} & 0.12 & \colorbox{correctPredictionColor}{1.0} & 0.36 \\

\midrule
$\datasetLMOrderStaticWikipediaCVS$ & $\modelGemmaFourEtwoBInstruct$ & $\sdmHR$ & \colorbox{correctPredictionColor}{\allRejected} & 0.0 & \colorbox{correctPredictionColor}{\allRejected} & 0.0 & \colorbox{correctPredictionColor}{\allRejected} & 0.0 & \colorbox{correctPredictionColor}{\allRejected} & 0.0 & \colorbox{correctPredictionColor}{\allRejected} & 0.0 \\
$\datasetLMOrderStaticWikipediaCVS$ & $\modelGemmaFourEtwoBInstructFinetuned$ & $\sdmHR$ & \colorbox{correctPredictionColor}{1.0} & 0.10 & \colorbox{correctPredictionColor}{\allRejected} & 0.0 & \colorbox{correctPredictionColor}{1.0} & 0.10 & \colorbox{correctPredictionColor}{\allRejected} & 0.0 & \colorbox{correctPredictionColor}{1.0} & 0.10 \\
$\datasetLMOrderStaticWikipediaCVS$ & $\modelGemmaFourEtwoBInstructFinetunedAdditionalRun$ & $\sdmHR$ & \colorbox{correctPredictionColor}{1.0} & 0.0027 & \colorbox{correctPredictionColor}{\allRejected} & 0.0 & \colorbox{correctPredictionColor}{1.0} & 0.0027 & \colorbox{correctPredictionColor}{\allRejected} & 0.0 & \colorbox{correctPredictionColor}{1.0} & 0.0027 \\
$\datasetLMOrderStaticWikipediaCVS$ & $\modelGemmaFourEtwoBInstructFinetunedCM$ & $\sdmHR$ & \colorbox{correctPredictionColor}{1.0} & 0.34 & \colorbox{correctPredictionColor}{1.0} & 0.21 & \colorbox{correctPredictionColor}{1.0} & 0.34 & \colorbox{correctPredictionColor}{1.0} & 0.21 & \colorbox{correctPredictionColor}{1.0} & 0.55 \\
$\datasetLMOrderStaticWikipediaCVS$ & $\modelGemmaFourEtwoBInstructFinetunedCMAdditionalRun$ & $\sdmHR$ & \colorbox{correctPredictionColor}{1.0} & 0.27 & \colorbox{correctPredictionColor}{1.0} & 0.13 & \colorbox{correctPredictionColor}{1.0} & 0.27 & \colorbox{correctPredictionColor}{1.0} & 0.13 & \colorbox{correctPredictionColor}{1.0} & 0.41 \\

    \bottomrule
  \end{tabular}
  }  
    \caption{Comparison of estimators trained and evaluated on \underline{static} positive and negative sequences, with \colorbox{correctPredictionColor}{$\alpha$}$=0.99$, averaged over 6 independently calibrated $\sdm$ estimators. \colorbox{correctPredictionColor}{\allRejected} indicates all predictions were rejected across all estimators, which is preferred over falling \colorbox{wrongPredictionColor}{under} the expected accuracy. (Highlighting is applied prior to rounding.)
  $n=|\text{Admitted}|$, the count of non-rejected documents. Unlike Table~\ref{tab:experiments-verification-results}, the evaluation sets contain the same static and positive sequences across all models.} %
  \label{tab:experiments-verification-results-static} 
\end{table*}

\end{document}